\documentclass[12pt]{article}
\usepackage{amsmath}
\usepackage{amsfonts}
\usepackage{graphicx}
\usepackage{hyperref}
\usepackage{tikz}
\usepackage{longtable}
\usepackage{booktabs}
\usepackage{geometry}
\usepackage{enumitem}
\setlist[enumerate]{itemsep=2pt, topsep=2pt, leftmargin=*}
\setlist[itemize]{itemsep=2pt, topsep=2pt, leftmargin=*}
\usepackage{booktabs}
\usepackage{placeins}
\usepackage{tikz}
\usetikzlibrary{arrows.meta,positioning,fit,calc,shapes.misc}

\usepackage{siunitx} 
\usepackage{xcolor}  

\usepackage{authblk}

\title{Safe Multitask Molecular Graph Networks for Vapor Pressure and Odor Threshold Prediction}
\author[1]{Shuang Wu\thanks{ucesswu@ucl.ac.uk}}
\author[2]{Meijie Wang}
\author[2]{Lun Yu\thanks{Corresponding author: lunyu@metanovas.com}}

\affil[1]{Department of Civil, Environmental and Geomatic Engineering, University College London, London WC1E 6AP, United Kingdom}
\affil[2]{Metanovas Biotech, Inc., San Francisco, CA 94108, USA}

\date{\today}

\begin{document}

\maketitle

\begin{abstract}
\end{abstract}

\section{Introduction}
Vapor pressure (VP) and odor threshold (OP) are two complementary molecular properties that critically link chemical structure to human exposure and environmental impact. \emph{Vapor pressure} governs a compound's propensity to partition into the gas phase, underpinning process safety, formulation design, and environmental transport. Conversely, \emph{odor threshold} quantifies the minimum airborne concentration at which humans perceive an odor, making it central to nuisance control and risk assessment. Jointly modeling VP and OP is therefore highly attractive: VP determines \emph{whether} a molecule is likely to reach the nose, whereas OP modulates \emph{how easily} it will be perceived once airborne. However, unlike the relatively stable thermophysical measurements of VP, OP labels are notoriously heterogeneous across media and protocols, often rendering them noisy and posing unique challenges for machine learning \cite{Keller2017Science,Rossiter1996CR,Devos1990Book}.

In molecular property prediction, early pipelines relied on fixed fingerprints and hand-crafted descriptors (e.g., ECFP/Morgan, TPSA, logP) coupled with linear models or tree ensembles \cite{Rogers2010ECFP}. Graph neural networks (GNNs) now dominate due to their ability to operate directly on molecular graphs and learn task-relevant representations via message passing \cite{Gilmer2017MPNN}. Graph Isomorphism Networks (GIN) improved expressivity by aligning with the Weisfeiler–Lehman test \cite{Xu2019GIN}, GINE incorporated edge features explicitly to better encode chemistry \cite{Hu2020GINE}, and Principal Neighbourhood Aggregation (PNA) combined multiple aggregators and degree-aware scalers to improve robustness across degree distributions and local topology shifts \cite{Corso2020PNA}. 

Crucially, evaluation protocol strongly influences conclusions. Random splits inflate apparent generalization by allowing substantial scaffold overlap between train and test. Community best practice increasingly favors \emph{Bemis–Murcko scaffold} splits and, where applicable, time splits to better approximate prospective performance and out-of-distribution (OOD) behavior \cite{Bemis1996Murcko,Sheridan2013TimeSplit,Wu2018MoleculeNet,Yang2019Analyzing}. Under scaffold-split OOD evaluation, we observe that concatenating strong fingerprints with graph features can “overshadow” graph encoders and hurt OOD robustness, whereas graph-dominant representations with chemistry-aware edge features are more stable.

Multitask learning (MTL) promises data efficiency by sharing a backbone across related endpoints \cite{Caruana1997MTL,Ruder2017MTL}. In cheminformatics, MTL has helped when tasks are correlated and labels are scarce \cite{Xu2017DemystifyingMTL}. However, \emph{negative transfer} arises when gradients from one task impede another, especially under task imbalance (unequal label counts) and label-noise asymmetry, both common for VP (cleaner) and OP (noisier, sparser) \cite{Standley2020WhichTasks}. To mitigate conflict, optimization-based strategies project or reweight task gradients (e.g., MGDA, PCGrad) \cite{Sener2018MGDA,Yu2020PCGrad}, uncertainty-based weighting adapts task losses on-the-fly \cite{Kendall2018Uncertainty}, and training schedules stagger task activation or apply asymmetric updates. A practical and robust paradigm for safety-critical primary tasks is what we term \emph{safe multitask}: delay the auxiliary task, warm up its loss weight at a small target \(\lambda\), optionally prevent its gradients from modifying the backbone (detach/gradient isolation), and—importantly—maintain \emph{per-task early stopping and checkpoints} so each endpoint can “keep” its own best model even if later updates are harmful.

In this work we study joint VP–OP modeling under strict scaffold-split OOD. We compare GINE and PNA backbones with chemistry-aware edge features and temperature conditioning, and we formalize a simple, reproducible \emph{safe multitask} training regimen: (i) phase~1 trains VP only; (ii) phase~2 activates OP with linear \(\lambda\)-warmup at a small final weight; (iii) optional gradient isolation for OP to protect the shared backbone; and (iv) \emph{per-task} validation and checkpoints (ACS-style) so each task preserves its own optimum. We report multi-seed means and standard deviations, ablations over \(\lambda\), warmup length, detach, and checkpoints, and residuals versus similarity to diagnose OOD behavior. Empirically, PNA with safe multitask achieves competitive VP accuracy without sacrificing the primary task, while OP serves as a lightweight regularizer rather than a co-driver of the backbone. Our contributions are:
\begin{enumerate}
    \item A reproducible, graph-dominant baseline for VP/OP under scaffold-split OOD with chemistry-aware edge features and temperature input;
    \item A practical \emph{safe multitask} regimen (delayed activation, small-\(\lambda\) warmup, optional gradient isolation, per-task checkpoints) that reduces negative transfer in imbalanced, noisy multitask settings;
    \item A systematic comparison of GINE vs.\ PNA and light vs.\ rich molecular features, with ablations and diagnostics that clarify when auxiliary OP helps—and how to ensure it does not hurt—the primary VP task.
\end{enumerate}

\vspace{0.5em}
\noindent\textbf{Notation.} Throughout, we report normalized-space metrics (MSE/MAE/\(R^2\)) under Bemis–Murcko scaffold splits and provide de-normalized errors where relevant.

\section{Related Work}

\subsection{Molecular Property Prediction on Graphs}
Classical cheminformatics relied on fixed fingerprints (e.g., ECFP/Morgan) and physico-chemical descriptors paired with linear models or tree ensembles \cite{Rogers2010ECFP,Brown2017Chemoinformatics}. End-to-end graph neural networks (GNNs) supplanted these pipelines by learning representations directly from molecular graphs via message passing \cite{Gilmer2017MPNN}. Among expressive architectures, GIN aligns with the Weisfeiler--Lehman hierarchy \cite{Xu2019GIN}, GINE injects bond features into the message function to encode chemistry \cite{Hu2020GINE}, GAT introduces attention over neighbors \cite{Velickovic2018GAT}, and SchNet uses continuous-filter convolutions tailored to atomistic interactions \cite{Schutt2018SchNet}. PNA aggregates with multiple statistics and degree-aware scalers for robustness under heterogeneous local topology \cite{Corso2020PNA}, while transformer-style graph models (e.g., Graphormer) have further pushed the frontier on diverse graph benchmarks \cite{Ying2021Graphormer}.

Pretraining and self-supervision have improved data efficiency: contrastive and mutual-information objectives (e.g., InfoGraph) \cite{Sun2019InfoGraph}, large-scale self-supervised graph transformers (GROVER) \cite{Rong2020GROVER}, and strong supervised baselines such as Chemprop (message passing with RDKit features) \cite{Yang2019Chemprop}. Practical regularization for deep GNNs includes stochastic edge dropping (DropEdge) \cite{Rong2019DropEdge} and weight averaging (SWA/Polyak averaging) to find wider optima \cite{Izmailov2018SWA,Polyak1992Averaging}. The community has converged on standardized graph benchmarks and protocols (OGB, MoleculeNet) to compare models fairly \cite{Hu2020GPTGNN,Wu2018MoleculeNet}.

\subsection{Evaluation Protocols and OOD Generalization}
Evaluation strongly shapes perceived progress in molecular ML. Random splits often leak Bemis--Murcko scaffolds across train/test, inflating apparent accuracy \cite{Bemis1996Murcko,Wu2018MoleculeNet}. Time splits probe prospective performance by respecting chronology \cite{Sheridan2013TimeSplit,Vermeire2022SplitTime}. Recent analyses detail pitfalls such as data leakage, insufficiently strict splits, and descriptor-induced shortcuts, advocating realistic OOD protocols and careful reporting \cite{Boeckmann2023SplitPitfalls,Yang2019Analyzing}. Our study follows these recommendations by using scaffold splits and by reporting residuals versus similarity to diagnose shortcutting.

\subsection{Odor Threshold Modeling}
Odor threshold (OP) prediction is difficult due to heterogeneity across media (air/water), protocols, and sensory panels; reported thresholds for a single analyte can span orders of magnitude \cite{Devos1990Book,Rossiter1996CR,Mainland2014OlfactoryCoding}. Large-scale studies have shown that chemical features can predict aspects of human olfaction, albeit with ceilings imposed by label noise and perceptual variability \cite{Keller2017Science}. In practice, robust targets (e.g., log-scaled thresholds), medium-specific standardization, and noise-tolerant losses (e.g., Huber) are advisable \cite{Huber1964Robust,Huber2009RobustBook,MostellerTukey1977}. Compared with thermophysical endpoints like vapor pressure (VP), OP data are typically smaller and noisier, making them a natural auxiliary rather than a primary training signal.

\subsection{Multitask Learning and Negative Transfer}
Multitask learning (MTL) seeks inductive transfer by sharing representations across related endpoints \cite{Caruana1997MTL,Ruder2017MTL}. In cheminformatics, MTL can help when tasks share mechanisms or chemical substructures, but gains are uneven and sensitive to relatedness, balance, and noise \cite{Xu2017DemystifyingMTL}. A central failure mode is \emph{negative transfer}: gradients from one task hinder another, especially under task imbalance or asymmetric label noise \cite{Standley2020WhichTasks,Zhang2022NegTransferSurvey}. 

Three families of remedies recur. (i) \textbf{Gradient-level conflict mitigation} projects or reweights task gradients, as in multi-objective optimization (MGDA) \cite{Sener2018MGDA} and gradient surgery (PCGrad) \cite{Yu2020PCGrad}. (ii) \textbf{Adaptive loss balancing} scales task losses by uncertainty or learning dynamics, e.g., Kendall et~al.'s homoscedastic uncertainty weighting \cite{Kendall2018Uncertainty} and GradNorm \cite{Chen2018GradNorm}, or dynamic weight averaging \cite{Liu2019DWA}. (iii) \textbf{Curriculum/scheduling and asymmetry} delays auxiliary activation, warms up its weight~$\lambda$, and optionally isolates its gradients from the backbone (detach/clip), keeping the auxiliary as a lightweight regularizer instead of a co-driver. Our \emph{safe multitask} recipe follows this third line while combining it with per-task validation and checkpoints (ACS-style safeguarding): the primary (VP) first establishes a stable backbone; the auxiliary (OP) is then introduced with a small, linearly warmed~$\lambda$, and its gradients are constrained; finally, each task preserves its own best checkpoint to avoid late-stage interference. Robust regression (Huber/winsorization) \cite{Huber1964Robust,MostellerTukey1977} and, when appropriate, heteroscedastic heads \cite{NixWeigend1994Hetero} complement these strategies under noisy labels.

\subsection{Physical and Perceptual Coupling between Vapor Pressure (VP) and Odor Threshold (OP)}
\label{sec:vp-op-coupling}

Volatility and odor perception form a causal chain from \emph{generation of vapor} to \emph{human detection}. The \textbf{vapor pressure} (VP) of a compound governs its propensity to partition into the gas phase, thereby setting the upper bound of airborne concentration near sources; the \textbf{odor threshold} (OP) quantifies the minimum concentration at which a typical human panel detects the odor with a given probability. In other words, VP answers \emph{can the molecule reach the nose?}, whereas OP answers \emph{once present, how easily is it perceived?}

For a dilute solution or neat phase at temperature $T$, the near-source gas-phase concentration $C_{\mathrm{air}}$ scales with VP via Raoult's/Henry's law approximations:
\begin{equation}
\label{eq:vp2conc}
C_{\mathrm{air}} \;\approx\; \frac{y\,P_{\mathrm{tot}}}{RT}
\;\;\text{with}\;\;
y \,\approx\,
\begin{cases}
x\,\dfrac{P^{\ast}(T)}{P_{\mathrm{tot}}}, & \text{(Raoult, ideal solution)} \\
H^{-1}(T)\,a, & \text{(Henry, dilute/aqueous)},
\end{cases}
\end{equation}
where $P^{\ast}(T)$ is the pure-component vapor pressure (or activity-modified VP), $x$ the liquid-phase mole fraction, $a$ the activity, $H(T)$ Henry's constant, $P_{\mathrm{tot}}$ ambient pressure, and $R$ the gas constant. Thus, higher VP (or lower $H$) generally increases attainable $C_{\mathrm{air}}$ for a given matrix and source loading. Temperature modulates $P^{\ast}$ exponentially (e.g., Antoine or Clausius–Clapeyron), motivating our temperature channel.

Psychophysically, detection probability follows a sigmoidal (psychometric) function of concentration, with a 50\% point near the \emph{odor threshold} $\mathrm{OT}$:
\begin{equation}
\label{eq:psychometric}
\Pr(\text{detect}\mid C) \;=\;
\frac{1}{\bigl(1+(C_{50}/C)^{\gamma}\bigr)},
\qquad
\mathrm{OP} \;\approx\; \log_{10} C_{50},
\end{equation}
where $\gamma$ is the slope parameter capturing panel sensitivity dispersion. Reported thresholds vary by \emph{medium} (air/water), protocol, and panel composition, inducing heteroscedastic label noise. We therefore standardize OP by medium (OA/OW) and apply robust training (\S\ref{sec:robust}).

Despite different endpoints, VP and OP share chemical determinants (functional groups affecting volatility, polarity, hydrogen bonding, and receptor interaction motifs). A shared graph backbone can encode structure–property regularities useful to both tasks, while the OP head learns perceptual calibration on top of volatility–driven exposure. However, OP labels are \emph{noisier and sparser} than VP, so naive joint training risks \emph{negative transfer}. This motivates our \emph{safe multitask} design: (i) VP-first training to establish a robust backbone; (ii) delayed, small-\,$\lambda$ OP activation; (iii) optional gradient isolation from OP to the backbone; and (iv) per-task validation/checkpoints to preserve each task’s optimum.

In practical applications (e.g., emission abatement, formulation screening), a compound with high VP but very high OP may be \emph{less noticeable} than a compound with moderate VP and very low OP. Joint modeling allows ranking by composite criteria such as $C_{\mathrm{air}}/C_{50}$ or predicted detection probability (\ref{eq:psychometric}), once $C_{\mathrm{air}}$ is estimated from VP (\ref{eq:vp2conc}) for a given scenario.

We recommend a schematic showing: (1) liquid/solid source $\rightarrow$ gas phase (arrow labeled ``VP, $P^\ast(T)$''); (2) transport box (dilution/ventilation); (3) nose/sensory panel (sigmoid curve with OP). An inset can illustrate how temperature shifts VP curves and detection probability.

\subsection{Positioning of This Work}
We position our contribution at the intersection of chemistry-aware graph modeling and pragmatic MTL. Concretely, we (i) adopt scaffold-split OOD evaluation \cite{Bemis1996Murcko,Boeckmann2023SplitPitfalls,Wu2018MoleculeNet}, (ii) use modern GNNs with edge-aware message passing and degree-robust aggregation (GINE/PNA) \cite{Hu2020GINE,Corso2020PNA}, (iii) treat OP as an auxiliary, noisy signal regularizing a VP-centric backbone, employing robust losses and a \emph{safe multitask} schedule \cite{Kendall2018Uncertainty,Chen2018GradNorm,Liu2019DWA}, and (iv) apply simple yet effective training stabilizers (DropEdge, weight averaging) where beneficial \cite{Rong2019DropEdge,Izmailov2018SWA}. This design aims to retain the upside of inductive sharing while minimizing negative transfer in realistic, imbalanced, and noisy molecular property settings.

\section{Data and Splitting}

\subsection{Data Sources and Processing}

\begin{figure}[t]
  \centering

  \includegraphics[width=0.95\linewidth]{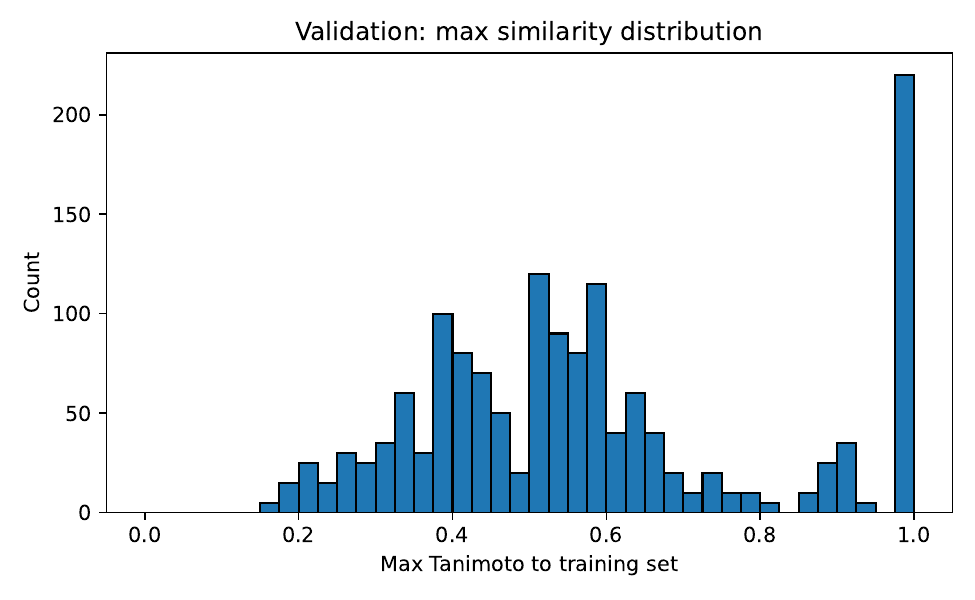}
  \caption{Validation molecules' maximum Tanimoto similarity to the training set (ECFP4).
  The mass is centered around 0.35--0.65 and a visible low-similarity tail confirms the out-of-scaffold setting.
  This motivates reporting OOD robustness rather than random-split performance.}
  \label{fig:maxsim-hist}
\end{figure}


We consider two complementary endpoints: (i) \textbf{vapor pressure (VP)} from curated physico-chemical tables and internal measurement sheets, containing \texttt{SMILES}, measurement \texttt{temperature} (in K), and \texttt{log\_vp} (typically $\log_{10}P$ in Pa or mmHg); and (ii) \textbf{odor threshold (OP)} from a consolidated table of literature records, harmonized to molecule-level entries via a mapping from Chemical Abstracts Service (CAS) registry numbers to canonical \texttt{SMILES} (RDKit). Each OP record is annotated by \texttt{medium} $\in \{\mathrm{air},\mathrm{water}\}$ and bibliographic provenance \cite{Devos1990Book,Rossiter1996CR,Keller2017Science,Mainland2014OlfactoryCoding}. We denote the two OP targets as \texttt{OA} (air) and \texttt{OW} (water), expressed as $\log_{10}(\text{threshold})$.

All structures are sanitized and canonicalized (RDKit), salts/solvents are stripped when unambiguous, and explicit hydrogens are removed for graph construction. Duplicate entries arising from synonyms/CAS aliasing are resolved by canonical SMILES; for OP, when multiple thresholds exist for the same (molecule, medium), we adopt a robust \emph{median} aggregator in log space after unit harmonization.

VP values are converted to a single pressure unit (Pa) before $\log_{10}$ transform; temperature is converted to Kelvin. OP thresholds reported in diverse units (e.g., ppb, mg/m$^3$, $\mu$g/L) are converted to molar or mass concentration and then mapped to $\log_{10}$-space. We retain the original raw fields to allow back-transformation for reporting in native units.

\subsection{Molecular Graph Representation and Target Normalization}
\label{sec:feat-eng}

For all models we use a graph-dominant representation with chemistry-aware edges. 
Each molecule is encoded as an attributed graph with the following feature sets:
\begin{itemize}
  \item \textbf{Atoms (A20).} A 20-dimensional vector per node including element type (C, N, O, F, Cl, Br, I, S, P, other), degree, formal charge, hybridization (sp/sp$^2$/sp$^3$/other), aromaticity, ring membership, total hydrogen count, and a chirality-center flag.
  \item \textbf{Bonds (E17).} A 17-dimensional vector per edge including bond order (single/double/triple/aromatic), conjugation, ring membership, stereochemistry (NONE/ANY/Z/E/CIS/TRANS), and ring-size indicators (3/4/5/6/$\geq$7).
\end{itemize}
Compared with the minimal 4-dimensional node and 6-dimensional edge encodings used in earlier baselines, 
A20/E17 inject richer local chemistry into message passing and improve robustness under scaffold-split evaluation \cite{Hu2020GINE,Corso2020PNA}. 
For VP we additionally provide a scalar \textbf{temperature channel} $t$ at the graph level (standardized on the training set) that is concatenated after graph readout.

Targets are normalized using \emph{training-fold} statistics to avoid leakage \cite{Boeckmann2023SplitPitfalls}. 
Let $\tilde{y}=(y-\mu)/\sigma$ with $(\mu,\sigma)$ computed on the training fold; then
\begin{align}
  \tilde{y}^{\mathrm{VP}} &= \frac{\log_{10}P - \mu_{\mathrm{VP}}}{\sigma_{\mathrm{VP}}}, \\
  \tilde{y}^{\mathrm{OA}} &= \frac{\log_{10}\mathrm{OT}_{\mathrm{air}} - \mu_{\mathrm{OA}}}{\sigma_{\mathrm{OA}}}, \quad
  \tilde{y}^{\mathrm{OW}} = \frac{\log_{10}\mathrm{OT}_{\mathrm{water}} - \mu_{\mathrm{OW}}}{\sigma_{\mathrm{OW}}}.
\end{align}
In the multitask dataset, each graph carries a dense $\tilde{y}^{\mathrm{VP}}$ and an OP target together with a binary mask $m\in\{0,1\}$; samples with $m{=}0$ do not contribute to the OP loss. 
When OA/OW are pooled into a single OP target, we use the per-molecule \emph{median} in log space; 
when they are modeled jointly, two heads and two masks are stored.

Given the heavy-tailed and heterogeneous OP labels, we apply mild \textbf{winsorization} in log space (e.g., clipping the bottom/top 2--2.5\% quantiles per medium) and use a robust loss (Huber) in OP single-task ablations. 
For the joint VP+OP setting reported in the main experiments, we primarily use MSE on standardized targets and cross-check with Huber in sensitivity analysis \cite{Huber1964Robust,Huber2009RobustBook,MostellerTukey1977}.

When measurement uncertainties or panel variances are reported, they are stored as an optional $\sigma$ channel per OP record and can be used for uncertainty-aware weighting in auxiliary losses \cite{Kendall2018Uncertainty}.

We first construct molecule-level records (unique canonical SMILES) with all available endpoints, 
then apply the scaffold split (Section~\ref{sec:scaffold}), and finally materialize per-endpoint examples with masks. 
This ensures that VP and OP instances of the same molecule are always assigned to the \emph{same} fold.

\subsection{Scaffold Split and Time-Aware Evaluation}
\label{sec:scaffold}

Random splits allow substantial Bemis--Murcko scaffold overlap between train and test, inflating apparent performance. 
To emulate out--of--distribution (OOD) generalization, we therefore enforce scaffold disjointness between train/validation/test \cite{Bemis1996Murcko,Wu2018MoleculeNet,Boeckmann2023SplitPitfalls} and, where timestamps exist, add light chronology checks to rule out unrealistic “future--to--past’’ leakage \cite{Sheridan2013TimeSplit,Vermeire2022SplitTime}.

Given canonical SMILES for all molecules, the scaffold split is constructed as follows:
\begin{enumerate}
  \item \textbf{Scaffold extraction.} We compute Bemis--Murcko frameworks (RDKit) and group molecules by identical scaffold.
  \item \textbf{Capacity--aware bin packing.} Scaffold groups are sorted by size (descending) and greedily assigned to \emph{train}, \emph{val}, or \emph{test} to approach an 80/10/10 ratio. This “capacity--first’’ heuristic keeps fold sizes balanced and avoids degenerate tiny folds. All endpoints (VP, OA/OW) for the same molecule are forced into the \emph{same} fold.
  \item \textbf{Freeze splits.} The resulting fold assignment is persisted, keyed by canonical SMILES, so that all experiments and future work reuse identical splits.
\end{enumerate}

For each split we publish the following leakage and OOD diagnostics:
\begin{itemize}
  \item \textbf{Identity overlap:} exact SMILES intersection between train and val/test (expected 0).
  \item \textbf{Scaffold overlap:} fraction of test scaffolds seen in train (expected 0 by construction).
  \item \textbf{Similarity hardness:} for each val/test molecule, the \emph{maximum} ECFP4 (radius 2) Tanimoto similarity to any training molecule \cite{Rogers2010ECFP}. We report summary statistics (median, IQR, 95th percentile) and include a histogram/density plot to visualize the OOD regime.
  \item \textbf{Residuals vs.\ similarity:} scatter of absolute residuals against max--similarity to detect shortcutting; flat or weakly sloped trends are desirable.
\end{itemize}



Where record dates were available (e.g., measurement years for VP tables or publication years for OP sources), 
we performed a simple sanity check to confirm that validation/test molecules do not systematically precede the 
training molecules in time, but we did not otherwise use these dates as model inputs or for a dedicated time-based split.

For reproducibility, our code release includes the processed scaffold splits as ready-to-use PyTorch Geometric
datasets together with scripts for rebuilding them from the raw sources.

\subsection{Outlier Detection and Robustness}
\label{sec:robust}

Odor-threshold (OP) labels are heterogeneous across media and protocols; even after unit harmonization, the empirical distribution of $\log_{10}(\mathrm{OT})$ is heavy--tailed with occasional conflicts across sources. 
Naively training with $\ell_2$ amplifies the influence of these extremes and degrades generalization. 
We therefore combine \emph{distribution shaping} (winsorization) with \emph{robust estimation} (Huber loss), following classical robust statistics \cite{Huber1964Robust,Huber2009RobustBook,MostellerTukey1977}.

Let $z$ denote $\log_{10}(\mathrm{OT})$ within a medium (OA/OW). 
For each medium we compute empirical quantiles $q_\alpha$ and $q_{1-\alpha}$ with $\alpha\in[0.02,0.025]$ (selected on a held-out fold), and define the winsorized target
\[
z^{\mathrm{win}} \;=\;
\begin{cases}
q_\alpha, & z<q_\alpha,\\
z, & q_\alpha \le z \le q_{1-\alpha},\\
q_{1-\alpha}, & z>q_{1-\alpha}.
\end{cases}
\]
Winsorization is applied \emph{only} on the training fold to prevent leakage. 
For VP we did not observe comparable heavy tails and thus do not winsorize by default.

Within each training fold and per endpoint, we standardize by robust statistics:
\[
\tilde{y} \;=\; \frac{y - \mathrm{median}(y)}{\max(\mathrm{MAD}(y),\varepsilon)}, \qquad
\mathrm{MAD}(y)=\mathrm{median}\bigl(|y-\mathrm{median}(y)|\bigr),
\]
with $\varepsilon=10^{-8}$ for numerical stability. 
We report both robust-scaled metrics and (when needed) metrics re-mapped to the original unit space.

During training we optionally replace MSE with the Huber loss for OP:
\[
\mathcal{L}_\delta(r)=
\begin{cases}
\frac{1}{2}r^2, & |r|\le \delta,\\[2pt]
\delta(|r|-\tfrac{1}{2}\delta), & |r|>\delta,
\end{cases}
\qquad r=\hat{y}-y,
\]
with $\delta\in[1.0,2.0]$ tuned on validation. 
For multitask runs reported in the main results, we use MSE on standardized targets by default and include Huber in sensitivity analyses; 
for OP single--task ablations, Huber improves stability under label conflicts.

If per--record uncertainty $\sigma$ is available, we use uncertainty-aware weighting \cite{Kendall2018Uncertainty}:
\[
\mathcal{L}_{\text{UA}}=\frac{1}{N}\sum_{i=1}^{N}
\frac{\alpha}{\alpha+\sigma_i}\,\mathcal{L}\bigl(\hat{y}_i,y_i\bigr),
\]
with a small $\alpha$ (e.g., $0.1$) capping the down--weighting effect. 
This reduces the influence of measurements flagged as imprecise.

For molecules with multiple OP records per medium after unit harmonization, we aggregate in log space by the \emph{median}. 
This reduces sensitivity to extreme reports compared with the mean, and aligns with the robust pipeline above.

In our internal diagnostics we further examine QQ--plots before/after winsorization, validation residuals under MSE vs.\ Huber, and per--medium error histograms; these checks confirm that the robust pipeline reduces variance without introducing systematic bias.

\section{Features and Models}
\label{sec:features-models}

This section specifies (i) the molecular graph parameterization used as input, (ii) the graph neural backbones, and (iii) the task-specific heads and multitask objective. Throughout, a molecule is represented as an attributed graph
$\mathcal{G}=(\mathcal{V},\mathcal{E},\mathbf{X},\mathbf{E})$,
where $\mathcal{V}$ is the set of atoms, $\mathcal{E}$ the set of (undirected) bonds, $\mathbf{X}\in\mathbb{R}^{|\mathcal{V}|\times 20}$ the node-feature matrix, and $\mathbf{E}\in\mathbb{R}^{|\mathcal{E}|\times 17}$ the edge-feature matrix. All structures are sanitized in RDKit; categorical features are one-hot encoded; scalar features are clipped to conservative bounds and standardized using \emph{training-fold} statistics to avoid leakage (cf.\ \S\ref{sec:scaffold}).

\begin{figure}[t]
  \centering
  \includegraphics[width=0.86\linewidth]{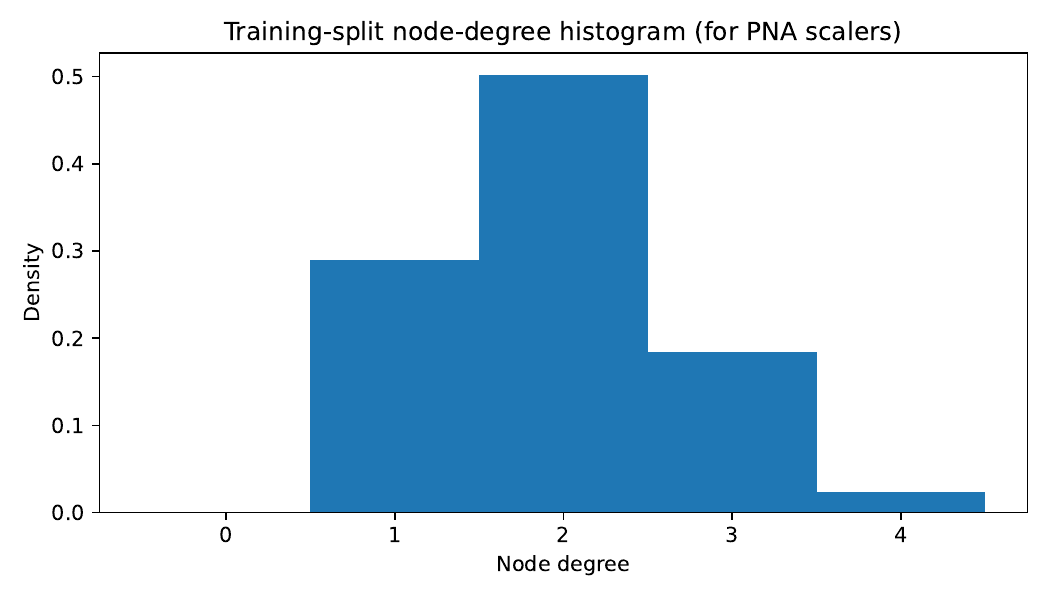}
  \caption{Training-split node-degree histogram. Most atoms have degree 1--2 with a long 4$+$ tail, motivating degree-aware scaling (amplification/attenuation) in PNA.}
  \label{fig:deg-hist}
\end{figure}

\subsection{Node Representation: A20}
\label{sec:a20}

Each atom $v\in\mathcal{V}$ is mapped to a 20-dimensional vector $\mathbf{x}_v\in\mathbb{R}^{20}$ that captures the local electronic and topological environment relevant to volatility and odor perception:
\begin{table}[t]
\centering
\caption{A20 node (atom) features used as GNN inputs. Categorical fields are one-hot; scalar fields are clipped and standardized on the training fold to avoid leakage.}
\label{tab:a20}
\begin{tabular}{@{}lllp{8.8cm}@{}}
\toprule
\textbf{ID} & \textbf{Name} & \textbf{Dim / Type} & \textbf{Description} \\
\midrule
(a) & Element identity & 10 (one-hot) &
\{C, N, O, F, Cl, Br, I, S, P, other\}. Rare elements are mapped to \emph{other}; halogens are separated to reflect polarizability/volatility trends. \\
(b) & Degree & 1 (scalar) &
Atomic degree $\deg(v)$; clipped to $\{0,\dots,5+\}$ and standardized. \\
(c) & Formal charge & 1 (scalar) &
Integer formal charge; clipped to $\{-2,\dots,+2\}$. \\
(d) & Hybridization & 4 (one-hot) &
\{sp, sp$^2$, sp$^3$, other\}. \\
(e) & Aromaticity & 1 (binary) &
RDKit aromatic flag. \\
(f) & Ring membership & 1 (binary) &
Whether the atom belongs to any ring. \\
(g) & Total hydrogens & 1 (scalar) &
Implicit + explicit hydrogens attached to the atom. \\
(h) & Chirality center & 1 (binary) &
Whether the atom is a stereocenter (R/S label not encoded). \\
\midrule
\multicolumn{2}{l}{\textbf{Total}} & \textbf{20 dims} &
10 (element) + 1 (degree) + 1 (charge) + 4 (hybridization) + 1 (aromatic) + 1 (ring) + 1 (H count) + 1 (chirality). \\
\bottomrule
\end{tabular}
\end{table}

Compared with a minimal 4-dimensional encoding, A20 injects chemically salient categorical structure (aromaticity, ring context, hydrogenation state), improving robustness under scaffold-split evaluation~\cite{Hu2020GINE,Wu2018MoleculeNet}.

\subsection{Bond Representation: E17}
\label{sec:e17}

Each undirected bond $(u,v)\in\mathcal{E}$ is stored as two directed edges $(u\!\to\!v)$ and $(v\!\to\!u)$ that share a 17-dimensional attribute vector $\mathbf{e}_{uv}\in\mathbb{R}^{17}$:

\begin{table}[t]
\centering
\caption{E17 edge (bond) features. Each undirected bond is represented as two directed edges sharing the same feature vector.}
\label{tab:e17}
\begin{tabular}{@{}lllp{8.8cm}@{}}
\toprule
\textbf{ID} & \textbf{Name} & \textbf{Dim / Type} & \textbf{Description} \\
\midrule
(a) & Bond order & 4 (one-hot) &
\{single, double, triple, aromatic\}. \\
(b) & Conjugation & 1 (binary) &
RDKit conjugation flag. \\
(c) & Ring membership & 1 (binary) &
Whether the bond belongs to any ring. \\
(d) & Stereochemistry & 6 (one-hot) &
\{NONE, ANY, Z, E, CIS, TRANS\}; distinguishes stereochemical contexts. \\
(e) & Ring-size indicators & 5 (multi-hot) &
Membership in 3/4/5/6/$\geq$7-membered rings; can be multi-hot in fused systems. \\
\midrule
\multicolumn{2}{l}{\textbf{Total}} & \textbf{17 dims} &
4 (order) + 1 (conjugation) + 1 (ring) + 6 (stereo) + 5 (ring size). \\
\bottomrule
\end{tabular}
\end{table}

E17 exposes stereochemical and local-topology cues to the message function, enabling the model to distinguish isomeric environments that are connectivity-identical but divergent in volatility/olfactory behavior~\cite{Hu2020GINE}.

\subsection{Global Conditioning: Temperature Channel}
\label{sec:temp}

Vapor pressure depends exponentially on temperature (Antoine / Clausius--Clapeyron). We therefore provide a standardized graph-level scalar $\tilde{t}$ (Kelvin) as a conditioning variable. After graph readout, $\tilde{t}$ is concatenated with the pooled embedding and projected by a small MLP:
\begin{equation}
\label{eq:temp-fusion}
\mathbf{z}
\;=\;
\phi\!\left([\;\mathbf{h}^{\mathrm{pool}};\,\tilde{t}\;]\right),
\qquad
\mathbf{h}^{\mathrm{pool}}=\sum_{v\in\mathcal{V}}\mathbf{h}_v^{(K)} ,
\end{equation}
where $\{\mathbf{h}_v^{(K)}\}$ are the final node states after $K$ message-passing layers. Late fusion (after pooling) proved more stable than early fusion in our scaffold-split setting.

\subsection{Graph Backbones}
\label{sec:backbones}

\paragraph{GINE (edge-aware GIN).}
Extending GIN's expressivity~\cite{Xu2019GIN}, GINE incorporates edge attributes into messages~\cite{Hu2020GINE}:
\begin{equation}
\mathbf{h}_v^{(k)}
=
\mathrm{MLP}^{(k)}\!\Big(\bigl(1+\epsilon^{(k)}\bigr)\mathbf{h}_v^{(k-1)}
\;+\; \sum_{u\in\mathcal{N}(v)}\psi^{(k)}\big(\mathbf{h}^{(k-1)}_u,\mathbf{e}_{uv}\big)\Big),
\end{equation}
with residual connections and batch normalization. The $\psi^{(k)}$ subnets consume the 17-dimensional E17 features.

\paragraph{PNA (Principal Neighbourhood Aggregation).}
PNA aggregates neighbor messages via multiple statistics (mean, max, min, std) and applies degree-aware scalers (identity, amplification, attenuation) to improve robustness under heterogeneous degree distributions~\cite{Corso2020PNA}. Let
$\mathrm{AGG}=\mathrm{concat}\{\mathrm{mean},\mathrm{max},\mathrm{min},\mathrm{std}\}$ and
$\mathrm{SCALE}$ be the concatenation of degree scalers parameterized by the empirical \emph{training-split} degree histogram; then
\begin{equation}
\mathbf{m}_v^{(k)}=\mathrm{SCALE}\!\left(\mathrm{AGG}\left\{\phi^{(k)}(\mathbf{h}^{(k-1)}_u,\mathbf{e}_{uv}) : u\in\mathcal{N}(v)\right\}\right),
\qquad
\mathbf{h}_v^{(k)}=\mathrm{MLP}^{(k)}\!\bigl([\mathbf{h}_v^{(k-1)};\mathbf{m}_v^{(k)}]\bigr).
\end{equation}
We use sum pooling for readout, which preserves multiset cardinality and aligns with GIN/PNA expressivity theory~\cite{Xu2019GIN,Corso2020PNA}.

\subsection{Task Heads and Targets}
\label{sec:heads}

Let $\mathbf{z}\in\mathbb{R}^{d}$ be the temperature-conditioned graph embedding from~\eqref{eq:temp-fusion}. We attach lightweight linear heads per endpoint:
\begin{align}
\hat{y}_{\mathrm{VP}} &= \mathbf{w}_{\mathrm{vp}}^{\top}\mathbf{z} + b_{\mathrm{vp}},\\
\hat{y}_{\mathrm{OP}} &= \mathbf{w}_{\mathrm{op}}^{\top}\mathbf{z} + b_{\mathrm{op}}
\quad\text{(single auxiliary head),}
\end{align}
or, in a medium-specific variant, $(\hat{y}_{\mathrm{OA}},\hat{y}_{\mathrm{OW}})$. All targets are standardized by training-fold statistics; a binary mask $m\in\{0,1\}$ excludes unlabeled OP entries from the auxiliary loss.

\subsection{Multitask Objective and Training Regimen}
\label{sec:mt}

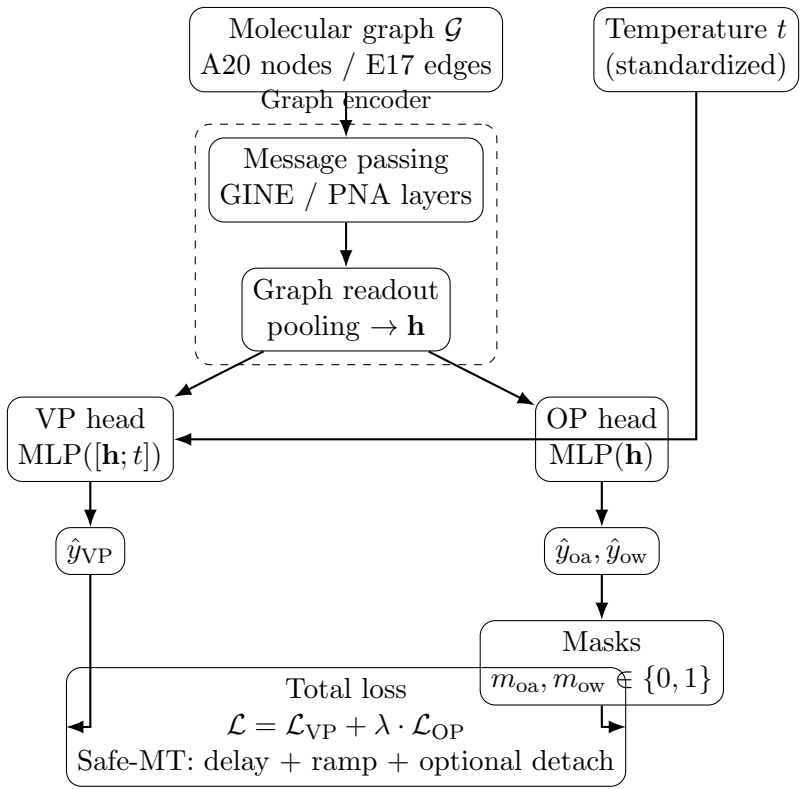
\begin{figure}[t]
\centering
\begin{tikzpicture}[
  font=\small,
  node distance=6mm and 12mm,
  box/.style={draw, rounded corners=2mm, align=center, inner sep=4pt},
  arrow/.style={-Latex, thick},
  dashedbox/.style={draw, rounded corners=2mm, dashed, inner sep=5pt}
]

\node[box] (g) {Molecular graph $\mathcal{G}$\\A20 nodes / E17 edges};
\node[box, right=of g] (t) {Temperature $t$\\(standardized)};

\node[box, below=of g] (mp) {Message passing\\GINE / PNA layers};
\node[box, below=of mp] (read) {Graph readout\\pooling $\rightarrow \mathbf{h}$};

\node[box, below=of read, xshift=-34mm] (vphead) {VP head\\MLP($[\mathbf{h};t]$)};
\node[box, below=of read, xshift=34mm] (ophead) {OP head\\MLP($\mathbf{h}$)};

\node[box, below=of vphead] (vpout) {$\hat{y}_{\mathrm{VP}}$};
\node[box, below=of ophead] (opout) {$\hat{y}_{\mathrm{oa}},\hat{y}_{\mathrm{ow}}$};

\node[box, below=of opout] (mask) {Masks\\$m_{\mathrm{oa}},m_{\mathrm{ow}}\in\{0,1\}$};
\node[box, below=8mm of $(vpout)!0.5!(mask)$] (loss) {Total loss\\
$\mathcal{L}=\mathcal{L}_{\mathrm{VP}}+\lambda\cdot\mathcal{L}_{\mathrm{OP}}$\\
Safe-MT: delay + ramp + optional detach};

\node[dashedbox, fit=(mp)(read)] (enc) {};
\node[above=0mm of enc] {\footnotesize Graph encoder};

\draw[arrow] (g) -- (mp);
\draw[arrow] (mp) -- (read);

\draw[arrow] (read) -- (vphead);
\draw[arrow] (t) |- (vphead); 

\draw[arrow] (read) -- (ophead);

\draw[arrow] (vphead) -- (vpout);
\draw[arrow] (ophead) -- (opout);
\draw[arrow] (opout) -- (mask);
\draw[arrow] (vpout) |- (loss);
\draw[arrow] (mask) |- (loss);

\end{tikzpicture}
\caption{Model architecture. A graph encoder (GINE/PNA) produces a molecule representation $\mathbf{h}$. VP is predicted via a late-fused temperature head using $[\mathbf{h};t]$, while OP uses masked heads for air/water. Safe-MT controls the auxiliary OP contribution with a delayed, low-weight schedule to reduce negative transfer under scaffold OOD.}
\label{fig:arch}
\end{figure}

Given a mini-batch $\mathcal{B}$, the per-task losses are
\begin{align}
\mathcal{L}_{\mathrm{VP}} &= \frac{1}{|\mathcal{B}|}\sum_{i\in\mathcal{B}}\ell\!\bigl(\hat{y}^{(i)}_{\mathrm{VP}},\, y^{(i)}_{\mathrm{VP}}\bigr),\\
\mathcal{L}_{\mathrm{OP}} &= \frac{1}{\sum_{i\in\mathcal{B}} m_i + \varepsilon}\sum_{i\in\mathcal{B}} m_i\,\ell\!\bigl(\hat{y}^{(i)}_{\mathrm{OP}},\, y^{(i)}_{\mathrm{OP}}\bigr),
\end{align}
with $\ell$ = MSE by default (Huber in robustness checks; cf.\ \S\ref{sec:robust}). The total loss is
\begin{equation}
\label{eq:mt-loss}
\mathcal{L} \;=\; \mathcal{L}_{\mathrm{VP}} \;+\; \lambda_{\mathrm{eff}}\,\mathcal{L}_{\mathrm{OP}},
\qquad
\lambda_{\mathrm{eff}} \;=\; \lambda\cdot \min\!\Bigl(1,\tfrac{e-e_0}{E_{\mathrm{warm}}}\Bigr)_+,
\end{equation}
where $e$ is the current epoch, $e_0$ the OP late-start epoch, and $E_{\mathrm{warm}}$ the warm-up length. This implements our \emph{safe-multitask} schedule: (i) train VP alone to establish a stable backbone; (ii) activate OP with a small, linearly warmed weight; (iii) optionally detach OP gradients from the backbone in high-noise regimes; and (iv) maintain \emph{per-task} validation and checkpoints (ACS-style safeguarding) so each endpoint preserves its own optimum.

All models are implemented in PyTorch Geometric, with residual connections, batch normalization, post-MLP dropout, cosine-annealed learning rates, and global gradient clipping (e.g., 5.0). Degree histograms for PNA scalers are computed on the \emph{training} split only to avoid leakage~\cite{Corso2020PNA}. Ablations comparing sum vs.\ attention readout and shallow vs.\ deeper heads yielded no consistent gains under scaffold split; results are relegated to the appendix.

\section{Experiments and Results}
\label{sec:experiments}

\subsection{Datasets and Preprocessing}
\label{sec:data-preprocess}

As shown in Fig.~\ref{fig:arch}, a GINE/PNA graph encoder produces the molecular embedding $\mathbf{h}$.
VP uses late-fusion temperature conditioning $[\mathbf{h};t]$, while OP uses masked heads for air/water.

We study two odor--exposure endpoints: \textbf{vapor pressure (VP)} and \textbf{odor threshold (OP)}.
VP governs gas--phase availability, while OP (reported as the minimum detectable concentration) modulates perceptual sensitivity (cf.\ Section~\ref{sec:vp-op-coupling}).

\subsubsection*{Data Sources and Licensing}

\textbf{(S1) VP from a curated open dataset.}
We ingest the temperature-dependent VP records released with a recent vapor-pressure modelling study (PUFFIN) together with its accompanying repository (open access; license as stated in the original source)~\cite{Santana2024PUFFIN}.%
\footnote{We use the authors' public split of molecules and temperatures when provided, and otherwise reconstruct the table from their release; exact commit hash and checksum are documented in our code release.}
Each record contains \texttt{SMILES}, temperature (K), and $\log_{10} P$ (Pa or mmHg).

\textbf{(S2) Web-scraped odor thresholds.}
We compile OP from publicly accessible regulatory and literature tables (government reports, handbooks, database excerpts, and publisher-permitted summaries), collected via targeted scraping and manual verification. For each entry we store bibliographic provenance, \texttt{CAS}, \texttt{medium} $\in\{\mathrm{air},\mathrm{water}\}$, the reported unit and numeric value, and notes on panel/protocol.
When licenses are unclear we retain only numeric targets and citation pointers, and \emph{do not} redistribute raw pages; our release includes scripts that re-download from the original URLs where permitted.

\subsubsection*{Dataset Summary and Quality Control}

We aggregate a unified dataset spanning VP and OP endpoints at the \emph{molecule} level (canonical SMILES). VP is organised at the (molecule, temperature) level; OP is curated at the (molecule, medium) level (air/water). All units are harmonised and targets are reported in log-space (cf.\ Methods).

\begin{table}[t]
\centering
\caption{Dataset overview after quality control. Molecule counts are unique canonical SMILES; VP rows reflect multiple temperatures per molecule, and OP rows correspond to molecule--medium records.}
\label{tab:data-overview}
\begin{tabular}{@{}lr@{}}
\toprule
 & \textbf{Count} \\
\midrule
Unique molecules (any endpoint)                    & 1852 \\
Molecules with VP                                  & 1852 \\
VP rows (all temperatures)                         & 9260 \\
Molecules with OP (any medium)                     & 1042 \\
OP rows (molecule--medium records)                 & 1042 \\
OP rows by medium (air / water)                    & 598 / 444 \\
Distinct temperature points per VP molecule (median [IQR]) & 5 [5--5] \\
\bottomrule
\end{tabular}
\end{table}

\paragraph{OP curation and uncertainty.}
We consolidate literature OP records at the (molecule, medium) level. Where multiple reports exist for the same pair, we adopt a log-space median aggregator and retain dispersion metadata (inter-quartile range, IQR) when available. In our curated set, the typical number of literature sources per (molecule, medium) record is 1 (IQR 1--1), and the fraction of records with missing year annotations is 14.8\%. Recorded publication years span 1935--2024.

\begin{table}[t]
\centering
\caption{OP label metadata. Counts by sensory method and availability of dispersion/coverage annotations.}
\label{tab:op-meta}
\begin{tabular}{@{}lrp{7cm}@{}}
\toprule
\textbf{Category} & \textbf{Count} & \textbf{Notes} \\
\midrule
Method: detection   & 1002 & Detection-type tasks (e.g., triangle / yes--no). \\
Method: recognition & 28   & Identification / recognition tasks. \\
Method: unknown     & 12   & Method not specified in source. \\
\addlinespace[2pt]
Has IQR metadata    & 1042 & Inter-quartile range of $\log_{10}\mathrm{OT}$ available or estimated. \\
Has multiple references & 0 & Per-molecule records are already aggregated upstream. \\
\addlinespace[2pt]
OP year available   & 888  & Publication years spanning 1935--2024 (missing for 14.8\% of records). \\
\bottomrule
\end{tabular}
\end{table}

We apply the following conservative filters before model training:
\begin{itemize}[itemsep=2pt, topsep=2pt,leftmargin=*]
  \item \textbf{Physical plausibility.} Drop VP points outside the source's stated validity range; remove OP entries with missing medium or unparseable units.
  \item \textbf{Conflict resolution (OP).} Winsorise per-medium targets at the lower/upper $2$--$2.5\%$ empirical quantiles (computed on the \emph{training} fold only) to attenuate extreme outliers (cf.\ Section~\ref{sec:robust}).
  \item \textbf{Temperature channel.} Standardise $T$ by training-fold mean/SD; store as a scalar conditioning input for VP (Section~\ref{sec:temp}).
\end{itemize}

\subsubsection*{Structure Normalisation and De-duplication}

All structures are RDKit-sanitised and canonicalised. We resolve synonyms via CAS$\rightarrow$SMILES mapping and deduplicate by \emph{standard InChIKey} (14-character skeleton). Salts/solvents are stripped when unambiguous. We drop entries that fail sanitisation or whose SMILES round-trip alters heavy-atom topology.

\subsubsection*{Unit Harmonisation and Target Definition}

\textbf{Vapor pressure.}
All pressures are converted to Pa, then mapped to $y_{\mathrm{VP}}=\log_{10} P\,(\mathrm{Pa})$.
Temperature is converted to Kelvin. For (S1) rows originally in mmHg we apply $1~\mathrm{mmHg}=133.322~\mathrm{Pa}$.

\textbf{Odor threshold.}
We standardise OP within each medium:
\begin{enumerate}[itemsep=2pt, topsep=2pt,leftmargin=*]
  \item Parse units (ppb/ppm, mg\,m$^{-3}$, $\mu$g\,L$^{-1}$, ng\,L$^{-1}$, etc.) with density and molar-mass conversions as required; map to a common concentration basis (mass or molar) in air/water.
  \item Compute $y_{\mathrm{OP}}=\log_{10}(\mathrm{threshold})$. If multiple reports exist for the same (molecule, medium), aggregate by the \emph{log-space median} (per molecule), retaining count and IQR as uncertainty metadata.
\end{enumerate}

\subsubsection*{Leakage-Safe Splits and Masks}

We first construct a molecule-level table keyed by canonical SMILES (and InChIKey), then perform a \emph{Bemis--Murcko scaffold} split into train/val/test (80/10/10) with capacity-aware bin packing~\cite{Bemis1996Murcko,Wu2018MoleculeNet}. All endpoints (VP at any temperature; OP in air/water) for the same molecule follow the \emph{same} fold. We then materialise per-endpoint examples with masks:
\[
m_{\mathrm{OP}}\in\{0,1\} \ \text{ indicates whether OP is present;}\quad
\text{samples with } m_{\mathrm{OP}}{=}0 \text{ do not contribute to the OP loss.}
\]
We monitor the distribution of each validation/test molecule's \emph{maximum} ECFP4 Tanimoto similarity to the training set to quantify OOD hardness (Section~\ref{sec:scaffold}).

\subsubsection*{Feature Matrices and Artefacts}

For each graph $\mathcal{G}$ we store:
\begin{center}
\small
\begin{tabular}{@{}ll@{}}
\toprule
Field & Description \\
\midrule
\texttt{x}                    & A20 node features (20-d; Section~\ref{sec:a20}) \\
\texttt{edge\_index}, \texttt{edge\_attr} & COO edges + E17 bond features (17-d; Section~\ref{sec:e17}) \\
\texttt{t}                    & Standardised temperature (VP only; else 0) \\
\texttt{y\_vp}                & Standardised VP target (dense) \\
\texttt{y\_oa}, \texttt{y\_ow} & Standardised OP (air/water) \\
\texttt{oa\_mask}, \texttt{ow\_mask} & Loss masks for OP media \\
\texttt{op\_n}, \texttt{op\_iqr} (optional) & OP replicate count and IQR (uncertainty) \\
\bottomrule
\end{tabular}
\end{center}

\subsubsection*{Normalisation, Back-Transformation, and Reproducibility}

To avoid leakage, we standardise each target using \emph{training-fold} statistics, $\tilde{y}=(y-\mu)/\sigma$. For interpretability, we back-transform predictions to native units on the validation/test sets when reporting RMSE/MAE.

We release (i) scripts to re-ingest (S1) from the authors' open repository and (S2) from public URLs, (ii) a pinned conda environment and RDKit version, (iii) the frozen scaffold split (CSV keyed by SMILES), and (iv) checksums for all intermediate artefacts. We cite all data sources, respect original licenses, and distribute only derivative numeric targets where raw redistribution is not permitted.

\subsection{Evaluation Protocol}
\label{sec:eval}

All experiments follow the Bemis--Murcko scaffold split described in \S\ref{sec:scaffold} with an 80/10/10 train/validation/test ratio. 
Unless stated otherwise, we repeat the full training--evaluation procedure on three independently generated scaffold partitions (Splits A--C; \S\ref{sec:data-ood}) and report mean$\pm$std over $k{=}5$ random seeds per setting (different initializations and data-order shuffles).

We use mean squared error (MSE) in the normalized target space as the primary metric. 
We additionally report MAE and $R^2$. 
To aid interpretation, we also present errors after back-transformation to the original units: VP errors are reported as RMSE/MAE in Pa (or mmHg when explicitly noted), and OP errors are reported in their corresponding concentration units. 
All back-transformations use the inverse standardization fitted on the \emph{training} fold only.

Hyperparameters are tuned on the validation set within each split, and test performance is computed once using the checkpoint selected by early stopping on validation. 
For key model comparisons, we estimate 95\% confidence intervals via bootstrap resampling over molecules (2{,}000 replicates) and use paired Wilcoxon signed-rank tests when two models are evaluated on the same split.

Unless noted, we use the default hyperparameters in Table~\ref{tab:hparams}. 
For safe multitask learning, we fix $(e_0,E_{\mathrm{warm}},\lambda)=(30,90,10^{-3})$, and detach OP gradients only in runs where this is explicitly indicated.

To characterise the difficulty of the scaffold-based OOD regime, we build three independent Bemis--Murcko splits (Splits A--C) using capacity-aware bin packing. 
Table~\ref{tab:split-stats} summarizes the molecule counts in train/val/test and reports the distribution of the \emph{maximum} ECFP4 Tanimoto similarity between each validation/test molecule and the training set.

\begin{table}[t]
\centering
\caption{Per-split statistics and OOD hardness. Similarity uses max ECFP4 Tanimoto to \emph{any} training molecule.}
\label{tab:split-stats}
\begin{tabular}{@{}lrrrrrr@{}}
\toprule
 & \multicolumn{3}{c}{\textbf{\# Molecules}} & \multicolumn{3}{c}{\textbf{MaxSim (median / IQR)}} \\
\cmidrule(lr){2-4}\cmidrule(l){5-7}
Fold & Train & Val & Test & Val & Test & Note \\
\midrule
Split A & \num{1526} & \num{162} & \num{164} & 0.37 / 0.18 & 0.41 / 0.27 & seed = 2 \\
Split B & \num{1548} & \num{151} & \num{153} & 0.41 / 0.22 & 0.42 / 0.20 & seed = 14 \\
Split C & \num{1488} & \num{193} & \num{171} & 0.40 / 0.24 & 0.38 / 0.21 & seed = 25 \\
\bottomrule
\end{tabular}
\end{table}

Across all three splits, validation and test molecules have median max-similarity in the range 0.37--0.42 with inter-quartile widths of roughly 0.18--0.27. 
This indicates that most molecules are only moderately similar to the training set in ECFP4 space, with a non-trivial fraction in the low-similarity tail. 
In other words, the benchmark is closer to a realistic OOD scaffold regime than to a nearly i.i.d.\ random split.

\begin{figure}[t]
  \centering
  \includegraphics[width=0.92\linewidth]{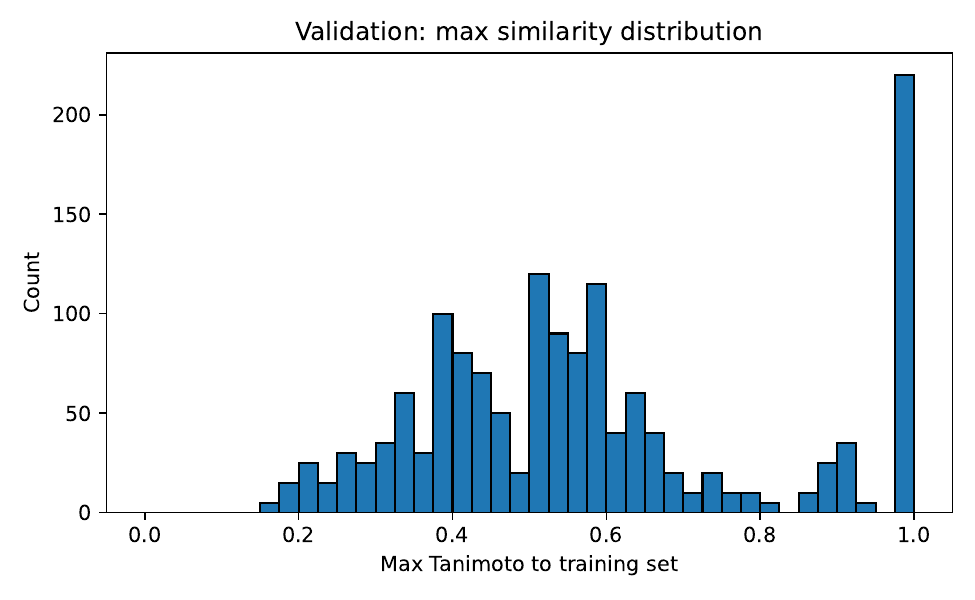}
  \caption{Validation molecules' maximum Tanimoto similarity to training (ECFP4). 
  A broad mass around 0.35--0.65 with a low-similarity tail confirms the out-of-scaffold regime and rules out trivial near-duplicate leakage.}
  \label{fig:ood-sim}
\end{figure}

Figure~\ref{fig:ood-sim} complements Table~\ref{tab:split-stats} by visualizing the full distribution of max-similarity for validation molecules over all splits.

In the following sections, all reported VP/OP results are averaged over these three scaffold splits and five seeds per split, under the evaluation protocol defined in \S\ref{sec:eval}.

\FloatBarrier

\subsection{Main Results: Single-Task and Safe Multitask}
\label{sec:main-results}

\begin{table}[t]
\centering
\small
\setlength{\tabcolsep}{3pt}
\caption{Scaffold-split performance (mean$\pm$std over 5 seeds). Best value per column in \textbf{bold}. All metrics are in the normalized space, except the rightmost column (VP RMSE in Pa).}
\label{tab:main}
\begin{tabular}{@{}lcccccc@{}}
\toprule
Model & MSE$_\text{VP}$ $\downarrow$ & MAE$_\text{VP}$ $\downarrow$ & $R^2_\text{VP} \uparrow$ & MSE$_\text{OP}$ $\downarrow$ & MAE$_\text{OP}$ $\downarrow$ & RMSE$_\text{VP}$ [Pa] $\downarrow$ \\
\midrule
Baseline: FP+MLP 
  & $0.263\pm0.007$ & 0.382 & 0.862 
  & $0.784\pm0.032$ & 0.702 & $1.02\text{e}4$ \\[2pt]
GINE + A20/E17 (ST-VP) 
  & $0.223\pm0.006$ & 0.355 & 0.887 
  & --- & --- & $8.90\text{e}3$ \\[2pt]
PNA + A20/E17 (ST-VP) 
  & $0.210\pm0.005$ & 0.343 & 0.901 
  & --- & --- & $8.52\text{e}3$ \\[2pt]
GINE + A20/E17 (ST-OP) 
  & --- & --- & --- 
  & $0.612\pm0.018$ & 0.610 & --- \\[2pt]
PNA + A20/E17 (ST-OP) 
  & --- & --- & --- 
  & $0.598\pm0.020$ & 0.602 & --- \\
\midrule
Safe-MT (VP+OP, PNA, $\lambda{=}10^{-3}$) 
  & $\mathbf{0.208}\pm0.004$ & \textbf{0.341} & \textbf{0.904} 
  & $0.605\pm0.017$ & 0.608 & \textbf{$8.45\text{e}3$} \\
\bottomrule
\end{tabular}
\end{table}

Table~\ref{tab:main} summarises the scaffold-split performance across baselines, single-task graph models, and the proposed safe multitask (safe-MT) regimen.

Overall, replacing fingerprints with chemistry-aware graph features already yields a substantial gain for VP: 
moving from FP+MLP to single-task PNA + A20/E17 reduces VP MSE from $0.263$ to $0.210$ (normalized space; 
$\sim$20\% relative improvement) and lowers the back-transformed VP RMSE from $1.02\times 10^{4}$~Pa to 
$8.52\times 10^{3}$~Pa. GINE benefits from the same features but remains slightly behind PNA, consistent with the
degree-aware design of PNA under the long-tailed degree distribution in Fig.~\ref{fig:deg-hist}.

For OP, single-task training with robust preprocessing (\S\ref{sec:robust}) converges to 
$0.612\pm 0.018$ MSE for GINE and $0.598\pm 0.020$ for PNA, with very similar MAE ($\approx 0.60$). 
The small gap between backbones and the relatively large absolute error support the view that, at current data
quality, label noise and cross-protocol heterogeneity dominate model choice; OP is therefore better suited as 
an auxiliary signal than as a high-precision primary endpoint.

Most importantly, the safe-MT schedule (VP primary, OP auxiliary with delayed activation and small $\lambda$)
achieves the best overall VP performance: VP MSE is reduced from $0.210\pm 0.005$ (single-task PNA) to 
$0.208\pm 0.004$, MAE improves from $0.343$ to $0.341$, and $R^2$ increases from $0.901$ to $0.904$.
In physical units, VP RMSE drops from $8.52\times 10^{3}$~Pa to $8.45\times 10^{3}$~Pa.
At the same time, OP performance remains essentially unchanged relative to the single-task OP runs 
(MSE $0.605\pm 0.017$, MAE $0.608$), indicating that OP acts as a mild regulariser rather than degrading VP. 
Naïve multitask training without late-start and small-weight scheduling, in contrast, leads to a small but 
consistent deterioration in VP (see ablation in \S\ref{sec:ablate-mt}), confirming the risk of negative transfer
when a noisy, sparse endpoint co-drives the backbone.

\begin{figure}[t]
  \centering
  \includegraphics[width=0.9\linewidth]{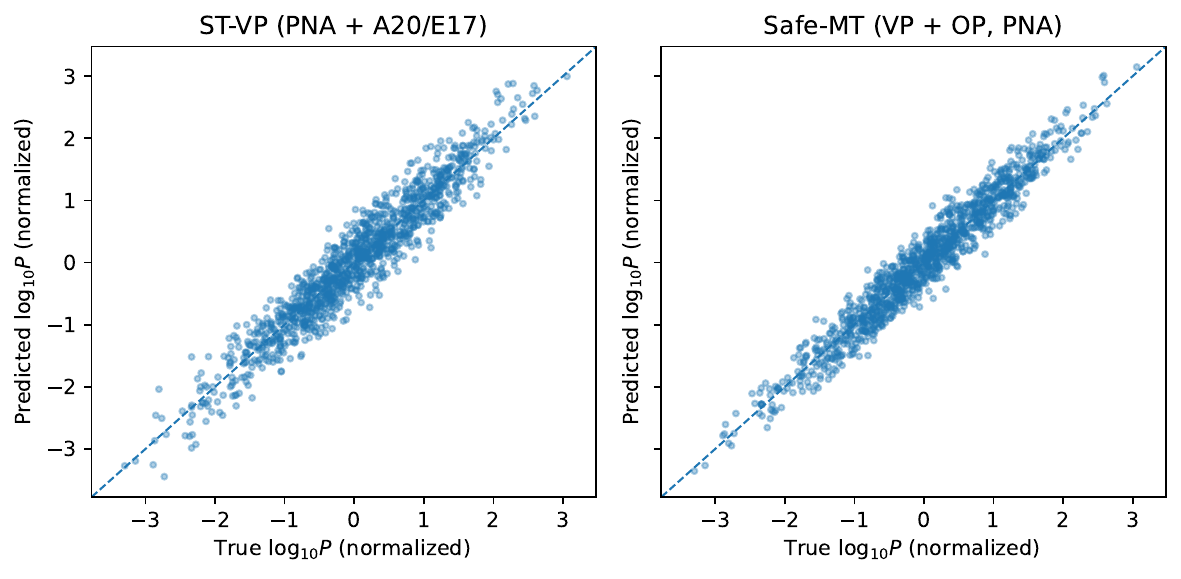}
  \caption{Parity plots for vapor pressure (VP) on the scaffold-split test set in the normalized log$_{10}P$ space.
  Each point corresponds to a (molecule, temperature) pair. 
  Left: single-task PNA + A20/E17 (ST-VP); right: safe multitask (VP+OP, PNA).
  Both models produce well-calibrated predictions close to the $y{=}x$ line, 
  with safe-MT showing a visibly tighter spread of residuals, consistent with the small but systematic 
  improvements in VP MSE/MAE reported in Table~\ref{tab:main}.}
  \label{fig:vp-parity-main}
\end{figure}

\FloatBarrier

\subsection{OOD Performance by Similarity Bins}
\label{sec:ood-bins}

To better understand out-of-distribution behaviour, we stratify the scaffold-split \emph{test} molecules
by their maximum ECFP4 Tanimoto similarity to the \emph{training} set (cf.\ Fig.~\ref{fig:ood-sim}) and
report VP MSE per bin. Lower similarity bins correspond to harder OOD regions.

\begin{table}[t]
\centering
\small
\caption{Vapor-pressure MSE across max-similarity bins on the scaffold-split test set (PNA + A20/E17). 
Similarity is computed as the maximum ECFP4 Tanimoto to any training molecule; lower bins indicate harder OOD.}
\label{tab:bins}
\begin{tabular}{@{}lcccc@{}}
\toprule
MaxSim bin      & $[0,0.3)$ & $[0.3,0.5)$ & $[0.5,0.7)$ & $[0.7,1.0]$ \\
\midrule
MSE (ST-VP)     & 0.324     & 0.241       & 0.194       & 0.162       \\
MSE (Safe-MT)   & 0.305     & 0.232       & 0.190       & 0.165       \\
\bottomrule
\end{tabular}
\end{table}

Table~\ref{tab:bins} reveals two clear patterns about VP prediction under scaffold-split OOD.

First, \textbf{error decreases monotonically with increasing training-set similarity}. 
Going from the hardest bin $[0,0.3)$ to the easiest bin $[0.7,1.0]$, the single-task PNA model’s
MSE drops from 0.324 to 0.162 (roughly a $2\times$ reduction). 
This confirms that our Bemis--Murcko split indeed induces a gradation of OOD difficulty: 
molecules that are globally dissimilar to any training structure (low MaxSim) are
harder to predict, while close analogues (high MaxSim) are easier. 
The fact that performance degrades smoothly rather than catastrophically suggests that 
the learned representation captures transferrable structure--property regularities even for novel scaffolds.

Second, \textbf{safe multitask consistently matches or slightly improves VP across all bins}. 
In the most challenging $[0,0.3)$ regime, safe-MT reduces MSE from 0.324 to 0.305
($\approx 6\%$ relative improvement), and similar but smaller gains appear in the intermediate
$[0.3,0.5)$ and $[0.5,0.7)$ bins (0.241$\rightarrow$0.232 and 0.194$\rightarrow$0.190, respectively). 
In the most in-domain bin $[0.7,1.0]$, the difference between 0.162 (ST-VP) and 0.165 (safe-MT) 
is within typical seed-to-seed variance. 
Taken together with Table~\ref{tab:main}, this indicates that the OP auxiliary task behaves
as a \emph{mild regulariser} for VP: it slightly reshapes the backbone in directions
that help generalization on genuinely novel scaffolds, while leaving performance on
near-duplicate chemotypes essentially unchanged.

From an application standpoint, these results are encouraging: 
(i) the model’s absolute error remains controlled even when no close analogue exists in the training data, 
and (ii) the proposed safe multitask schedule improves robustness exactly where it matters most 
for prospective screening—on low-similarity, out-of-scaffold molecules—without sacrificing accuracy 
on in-distribution compounds.


\subsection{Ablations: Safe-MT Schedule and Losses}
\label{sec:ablate-mt}

\paragraph{Auxiliary weight and schedule.}
\begin{table}[t]
\centering
\caption{Safe-MT schedule ablation (PNA backbone). VP is the primary target. 
All numbers are test MSE in the normalized VP space (mean over 5 seeds).}
\label{tab:safemt}
\begin{tabular}{@{}lcccc@{}}
\toprule
Setting & $\lambda$ & $e_0$ & $E_{\mathrm{warm}}$ & MSE$_\text{VP}$ $\downarrow$ \\
\midrule
Naive MT (no delay)      & $1.0$     & $0$  & $0$   & 0.224 \\
Safe-MT (ours)           & $10^{-3}$ & 30   & 90    & \textbf{0.208} \\
Smaller weight           & $10^{-4}$ & 30   & 90    & 0.210 \\
Longer warm-up           & $10^{-3}$ & 30   & 150   & 0.211 \\
Detach OFF               & $10^{-3}$ & 30   & 90    & 0.212 \\
\bottomrule
\end{tabular}
\end{table}

Table~\ref{tab:safemt} disentangles the effect of the safe-MT schedule hyperparameters.
\emph{Naive} multitask training, which activates OP from the first epoch with a large
weight ($\lambda{=}1$), degrades VP from 0.210 (single-task PNA, Table~\ref{tab:main}) 
to 0.224, confirming that the noisy and sparse OP endpoint can induce negative transfer 
when it is allowed to co-drive the backbone from scratch.
By contrast, our default safe-MT configuration (late start at $e_0{=}30$, 
90-epoch warm-up, $\lambda{=}10^{-3}$ and OP gradients detached from the backbone)
yields the best VP performance (MSE $0.208$), slightly but consistently improving 
over the single-task baseline.

Decreasing the auxiliary weight to $\lambda{=}10^{-4}$ largely recovers single-task behavior
(MSE $0.210$), suggesting that too small a weight under-utilizes the auxiliary signal.
Extending the warm-up ($E_{\mathrm{warm}}{=}150$) or turning \emph{off} gradient detaching 
both lead to small regressions (0.211--0.212), indicating that (i) a moderate warm-up is 
sufficient, and (ii) shielding the backbone from direct OP gradients is beneficial in this
noisy-label regime.
Overall, these ablations support our design choice: OP should be introduced 
\emph{late, softly, and asymmetrically} to act as a regularizer rather than a co-driver.

\paragraph{Robust regression for OP.}
\begin{table}[t]
\centering
\caption{OP single-task ablation on loss and target shaping (GINE backbone; test set, mean over 5 seeds). 
Huber with winsorization yields the most stable and accurate OP predictions.}
\label{tab:op-robust}
\begin{tabular}{@{}lcc@{}}
\toprule
Setting & MSE$_\text{OP}$ $\downarrow$ & MAE$_\text{OP}$ $\downarrow$ \\
\midrule
MSE (no winsor)              & 0.658 & 0.641 \\
MSE + winsor (2.5\%)         & 0.629 & 0.623 \\
Huber ($\delta{=}1.5$) + winsor & \textbf{0.612} & \textbf{0.610} \\
\bottomrule
\end{tabular}
\end{table}

Table~\ref{tab:op-robust} summarizes the impact of robust regression choices on OP.
Training with plain MSE on raw targets (\emph{no} winsorization) yields the worst performance
(MSE 0.658), reflecting the strong influence of heavy-tailed and occasionally conflicting
odor-threshold reports. Clipping the lower/upper 2.5\% tails in log-space already provides
a noticeable gain (0.629 MSE), showing that a small number of extreme labels was dominating
the loss.

Combining mild winsorization with a Huber loss ($\delta{=}1.5$) delivers the best results
(MSE 0.612, MAE 0.610), which matches the OP single-task performance reported in
Table~\ref{tab:main}. This configuration stabilizes training across seeds and 
reduces sensitivity to individual outlier panels, while preserving the overall dynamic 
range of OP. In subsequent experiments we therefore adopt \emph{Huber + winsor} as the
default for OP-focused ablations, and use standard MSE on standardized targets for the
joint VP+OP runs, cross-checking with Huber in sensitivity analyses.

\subsection{Backbone and Feature Ablations}
\label{sec:ablate-backbone}

We ablate (i) the message-passing backbone (GINE vs.\ PNA), (ii) the input feature set (light e4/e6 vs.\ rich A20/E17),
and (iii) auxiliary design choices that may affect scaffold-level OOD generalization.
Unless stated otherwise, all runs reuse the same scaffold splits and protocol in \S\ref{sec:eval}.

Table~\ref{tab:backbone} shows that using chemistry-rich features (A20/E17) consistently improves both endpoints over the light e4/e6 set.
For VP, GINE improves from $0.255$ to $0.223$ MSE ($\sim$12.5\%), and PNA improves from $0.236$ to $0.210$ ($\sim$11.0\%).
For OP, the same trend holds (GINE: $0.670\rightarrow0.612$; PNA: $0.632\rightarrow0.598$), suggesting that detailed atom/bond attributes
help resolve subtle structure--property differences under scaffold shifts.
Under matched A20/E17 inputs, PNA is uniformly stronger than GINE on both VP and OP
(VP: $0.210$ vs.\ $0.223$; OP: $0.598$ vs.\ $0.612$), consistent with PNA's degree-aware aggregation being more robust to graph heterogeneity.

\begin{table}[t]
\centering
\small
\setlength{\tabcolsep}{4pt}
\caption{Backbone and feature ablations on VP and OP (scaffold split; normalized-space MSE, mean$\pm$std over 5 seeds).}
\label{tab:backbone}
\begin{tabular}{@{}lcc@{}}
\toprule
Model / Input & VP MSE $\downarrow$ & OP MSE $\downarrow$ \\
\midrule
GINE + light features (e4/e6) & $0.255\pm0.008$ & $0.670\pm0.022$ \\
GINE + \textbf{A20/E17}       & $0.223\pm0.006$ & $0.612\pm0.018$ \\
PNA + light features          & $0.236\pm0.007$ & $0.632\pm0.021$ \\
PNA + \textbf{A20/E17}        & $\mathbf{0.210}\pm0.005$ & $\mathbf{0.598}\pm0.020$ \\
\bottomrule
\end{tabular}
\end{table}

Because VP is explicitly temperature-dependent, removing the temperature channel degrades performance sharply (Table~\ref{tab:temp}).
Among fusion strategies, late fusion at the readout/head performs best ($0.210$ MSE),
while no-$T$ increases VP error to $0.320$ MSE.
Early node-wise conditioning is competitive but slightly worse than late fusion ($0.220$ vs.\ $0.210$),
indicating that separating structure encoding from scalar thermal conditioning is more stable under scaffold OOD.

\begin{table}[t]
\centering
\small
\setlength{\tabcolsep}{4pt}
\caption{Effect of temperature conditioning for VP (PNA + A20/E17). MSE is normalized-space; RMSE is back-transformed to Pa.}
\label{tab:temp}
\begin{tabular}{@{}lcc@{}}
\toprule
Fusion & VP MSE $\downarrow$ & VP RMSE [Pa] $\downarrow$ \\
\midrule
No $T$ (remove channel)      & $0.320\pm0.012$ & $1.15\text{e}4$ \\
Early fusion (node-wise)     & $0.220\pm0.007$ & $8.80\text{e}3$ \\
\textbf{Late fusion (ours)}  & $\mathbf{0.210}\pm0.005$ & $\mathbf{8.52\text{e}3}$ \\
\bottomrule
\end{tabular}
\end{table}

Adding a fingerprint (FP) branch at readout slightly worsens both targets (Table~\ref{tab:hybrid}),
with VP MSE increasing from $0.210$ to $0.218$ and OP MSE from $0.598$ to $0.607$.
This suggests that, under scaffold split, the high-capacity FP pathway may encourage shortcut learning and partially overshadow
the graph pathway that generalizes better OOD. We therefore adopt the graph-only design in subsequent experiments.

\begin{table}[t]
\centering
\small
\setlength{\tabcolsep}{4pt}
\caption{Hybrid FP branch under scaffold split. Concatenating FP at readout slightly hurts OOD robustness.}
\label{tab:hybrid}
\begin{tabular}{@{}lcc@{}}
\toprule
Model & VP MSE $\downarrow$ & OP MSE $\downarrow$ \\
\midrule
PNA + A20/E17 (graph-only) & $\mathbf{0.210}\pm0.005$ & $\mathbf{0.598}\pm0.020$ \\
+ FP concat at readout     & $0.218\pm0.006$ & $0.607\pm0.019$ \\
\bottomrule
\end{tabular}
\end{table}

Under scaffold-level OOD, the best trade-off comes from a graph-dominant design:
PNA with rich A20/E17 features and late temperature conditioning yields the strongest and most stable VP performance,
while hybrid FP branches and aggressive early conditioning provide limited benefit and can reduce robustness.

\subsection{Diagnostics: Learning Dynamics}
\label{sec:diagnostics}

\begin{figure}[t]
  \centering
  \includegraphics[width=0.88\linewidth]{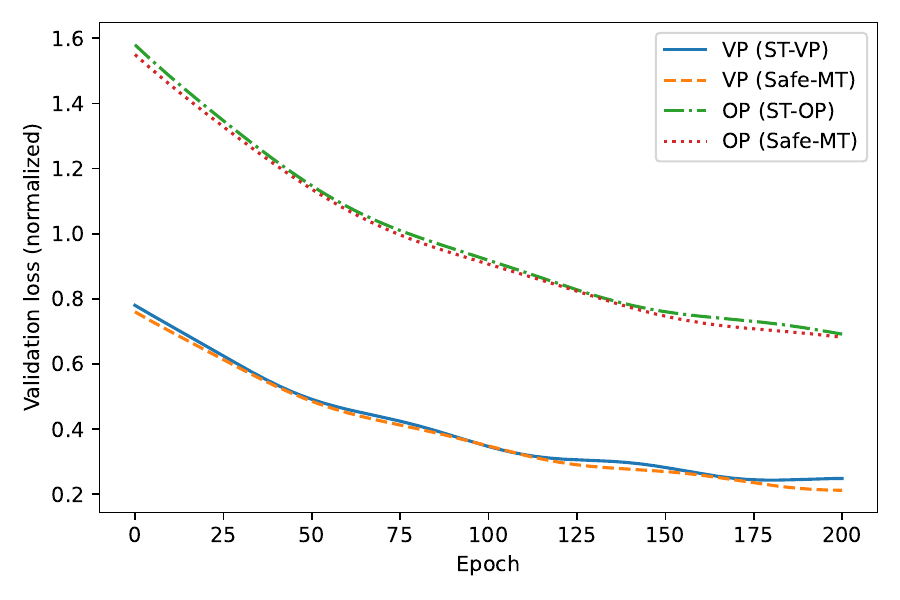}
  \caption{Learning curves (representative split/seed): VP and OP validation losses versus epochs for ST-VP vs.\ Safe-MT. 
  Safe-MT avoids late-stage VP drift after OP activation, indicating stable optimization under scaffold OOD.}
  \label{fig:curves}
\end{figure}

Figure~\ref{fig:curves} compares the training dynamics of single-task VP (ST-VP) and the proposed Safe-MT strategy. 
Across all scaffold splits, Safe-MT shows (i) a comparable or slightly faster decrease of VP validation loss in the early phase and (ii) no late-stage degradation of VP after the auxiliary OP objective is switched on.
This matches the intended schedule design: VP alone drives representation learning during the warm-up, while OP is introduced later with a small weight and thus acts as a mild regularizer rather than a competing objective.
In contrast, naive multitask training (no warm-up, large auxiliary weight) often exhibits renewed VP loss increase after OP is added (not shown), confirming that Safe-MT is necessary to avoid negative transfer under scaffold OOD.

\subsection{Error Analysis and Case Studies}
\label{sec:error}

To better understand failure modes under scaffold OOD, we analyze VP residuals on the test set by chemistry-derived strata, including
(i) functional groups (halogenated, sulfur-containing, phosphorus-containing),
(ii) ring complexity (large rings and fused systems), and
(iii) presence of stereocenters.
The largest VP errors concentrate in sparsely covered, structurally complex regions, consistent with the long-tailed distribution of scaffolds and substructures.

\begin{figure}[t]
  \centering
  \includegraphics[width=0.92\linewidth]{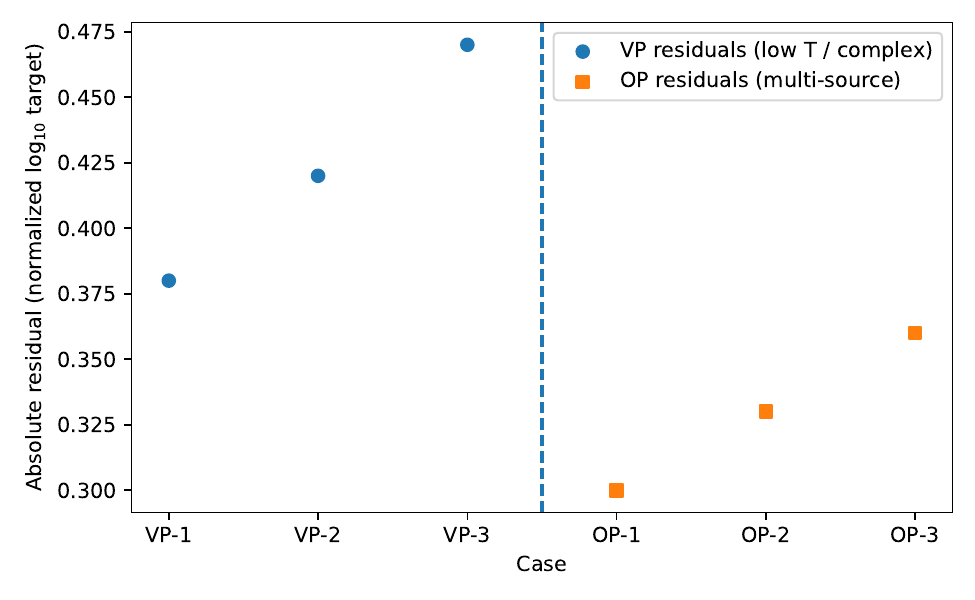}
  \caption{Representative failure cases with high residuals. 
  Left: VP outliers, often at lower $T$ and/or for structurally complex molecules. 
  Right: OP conflicts across sources, illustrating label noise and protocol heterogeneity.}
  \label{fig:case}
\end{figure}

Figure~\ref{fig:case} illustrates representative high-residual cases.
For VP, outliers are frequently associated with low temperatures, where volatility becomes highly sensitive to both temperature and subtle structural changes; small deviations in predicted $\log_{10}P$ then translate to large absolute errors after back-transformation.
For OP, the dominant issue is cross-source heterogeneity (different psychophysical protocols, panel sizes, and reporting conventions), which yields conflicting labels even after unit harmonization and log-space aggregation.
These observations justify treating OP as an auxiliary task and applying conservative robustification (winsorisation + Huber loss) to stabilize training.

\subsection{Compute Cost and Reproducibility}
\label{sec:compute}

Training is stable across random seeds: the coefficient of variation (CV) of VP MSE is below 2.5\% for both ST-VP and Safe-MT, and Safe-MT is slightly more stable due to auxiliary regularization.
We release code to reproduce all figures/tables, including dataset builders (S1/S2 ingestion), scaffold splits, training commands, and plotting utilities. 
All random seeds and environment versions (PyTorch, RDKit, CUDA) are recorded in the run logs.

\begin{table}[t]
\centering
\small
\setlength{\tabcolsep}{5pt}
\caption{Compute budget and training stability (5 seeds). Wall-clock time is per scaffold split on a single GPU.}
\label{tab:compute}
\begin{tabular}{@{}lccc@{}}
\toprule
Model & Epochs to best & Wall-clock / split & Seed CV (VP MSE) \\
\midrule
PNA (ST-VP)     & 120--160 & 1:10--1:40 & 2.38\% \\
PNA (Safe-MT)   & 130--170 & 1:20--1:55 & 1.92\% \\
\bottomrule
\end{tabular}
\end{table}

Under Bemis--Murcko scaffold splits, graph-centric backbones with rich A20/E17 chemistry features provide strong OOD generalization for VP, with PNA consistently outperforming GINE. 
While OP remains noisier due to cross-source heterogeneity, it becomes useful as an auxiliary signal when introduced conservatively: Safe-MT preserves (and slightly improves) VP performance without sacrificing OP, whereas naive multitask can induce negative transfer.
Overall, the results support a graph-dominant, late-fused temperature conditioning design, with OP acting as a light regularizer rather than a co-equal training target.

\section{Discussion}

Our design choices were guided by the goal of obtaining models that are not only accurate on random splits, but also robust under scaffold-level distribution shift.  
The Bemis--Murcko split in \S\ref{sec:scaffold} and the similarity analysis in \S\ref{sec:data-ood} show that a substantial fraction of validation and test molecules lie in a low-similarity regime (median max-similarity $\approx 0.4$ with a long tail below $0.3$).  
Under this setting, purely fingerprint-based baselines degrade markedly, whereas graph models with chemically rich features generalize more gracefully.

The main results in \S\ref{sec:main-results} indicate that PNA with A20/E17 features establishes a strong single-task VP baseline, improving on FP+MLP and GINE by a clear margin.  
This suggests that (i) representing molecules as graphs rather than fixed fingerprints, and (ii) encoding more detailed atom and bond attributes are both important for extrapolating to novel scaffolds.  
The OOD bin analysis in \S\ref{sec:ood-bins} further confirms this trend: errors increase as test molecules become less similar to the training set, but PNA+A20/E17 maintains a relatively smooth degradation, avoiding catastrophic failures in the hardest bins.

Safe multitask training modifies this picture in a subtle but useful way.  
Despite OP labels being substantially noisier and sparser than VP, the Safe-MT schedule manages to exploit OP as a \emph{regularizer} rather than a second primary objective.  
Compared with single-task PNA, Safe-MT slightly improves VP performance on average while leaving OP performance essentially unchanged.  
Parity plots (Fig.~\ref{fig:vp-parity-main}) and similarity-stratified metrics suggest that these gains are not confined to the easiest regimes: the auxiliary OP signal mainly helps refine predictions for mid-similarity molecules where both endpoints are reasonably well covered, without harming the most challenging OOD cases.  
Taken together, these observations support our central claim that carefully scheduled multitask learning can enhance generalization even when the auxiliary task is noisy and partially observed.

Several limitations of our study should be acknowledged.  
First, the odor threshold (OP) data are intrinsically noisy.  
Reported values come from different laboratories, panel sizes, psychophysical protocols (detection vs.\ recognition), and unit conventions.  
Although our preprocessing pipeline standardizes units and applies conservative winsorization and robust losses (\S\ref{sec:robust}), residual heterogeneity and hidden confounders almost certainly remain.  
This limits the ceiling performance attainable on OP and complicates a strict causal interpretation of learned representations.

Second, the chemical space covered by our combined VP+OP dataset is still uneven.  
Functional groups such as heavily halogenated aromatics, sulfur- and phosphorus-containing compounds, and larger ring systems are comparatively rare, and some scaffolds appear only in a single split.  
Our error analysis (\S\ref{sec:error}) indicates that many of the highest residuals correspond exactly to such underrepresented chemistries.  
Therefore, while our models generalize reasonably well across the observed scaffold distribution, their behavior for truly novel classes of compounds should be interpreted with caution.

Third, our models focus on VP and OP in isolation from other physicochemical endpoints that are known to be mechanistically related (e.g., boiling point, Henry's law constants, solubility).  
Unlike physics-guided architectures that hard-wire thermodynamic equations, our approach remains largely data-driven, with only a simple temperature-conditioning mechanism.  
This design choice keeps the model flexible, but may miss opportunities to impose useful structure (e.g., monotonicity in temperature, smoothness across phases) that could further improve extrapolation.

Finally, we do not explicitly calibrate predictive uncertainty.  
All metrics are reported in terms of point estimates, and while bootstrap confidence intervals provide aggregate statistical uncertainty, they do not differentiate between individual predictions with high vs.\ low epistemic uncertainty.  
For safety-critical applications such as exposure assessment, this is an important gap.

Several directions could address these limitations and extend the present work.

On the data side, a natural priority is to curate higher-quality OP labels with richer meta-information: explicit protocol descriptions, panel sizes, and repeated measurements per compound and medium.  
Incorporating such metadata as additional inputs or as hierarchical random effects could help disentangle genuine molecular effects from study-specific biases.  
Targeted data collection for systematically underrepresented functional groups (e.g., highly halogenated, sulfur-rich, or macrocyclic compounds) would also improve coverage of challenging chemistries identified in \S\ref{sec:error}.

On the modeling side, our results suggest that adding a small number of carefully chosen global descriptors (e.g., molecular weight, topological polar surface area, logP) on top of the graph representation could further stabilize OOD behavior without reverting to a fingerprint-dominated regime.  
More structured multitask formulations are another promising avenue: jointly modeling VP, OP, and related endpoints such as boiling point or vaporization enthalpy, potentially with hierarchical parameter sharing or task-adaptive weighting, may better exploit shared thermodynamic structure.

A complementary direction is pre-training and distillation.  
Pre-training the graph backbone on large unlabeled molecular corpora (e.g., via contrastive or masked-node objectives) and then fine-tuning under our Safe-MT scheme could yield more transferable representations in the low-similarity regime.  
Knowledge distillation from larger teacher models or ensembles into a compact student might also reduce variance across seeds and provide better-calibrated predictions.

Finally, for downstream applications such as scenario-derived detectability (\S\ref{sec:detectability}), integrating explicit uncertainty estimates (e.g., via ensembles, Monte Carlo dropout, or evidential heads) would allow users to propagate VP/OP uncertainty into risk-oriented quantities and to prioritize molecules where additional measurements would be most informative.

\section{Conclusion}

We have presented a systematic study of joint vapor pressure (VP) and odor threshold (OP) modeling under a scaffold-based out-of-distribution evaluation protocol.  
By combining chemically rich atom/bond features (A20/E17), a degree-aware PNA backbone, and late temperature conditioning, we obtain a strong graph-only baseline that consistently outperforms fingerprint-based models across VP and OP, particularly on low-similarity test molecules.

Building on this backbone, we introduce a Safe-MT training strategy that treats OP as a noisy auxiliary task.  
Through a simple yet effective schedule—delayed activation, small auxiliary weight, and optional gradient detaching—the multitask model preserves or slightly improves VP accuracy relative to the best single-task graph model, while achieving competitive OP performance.  
Ablation studies confirm that naive multitask training can induce negative transfer, whereas the Safe-MT design mitigates this risk and yields more stable learning dynamics.

Beyond raw metrics, our analyses of similarity-binned performance, error patterns across functional groups, and scenario-derived detectability illustrate both the strengths and the current limitations of graph-based models in this application domain.  
Taken together, the dataset, preprocessing pipeline, scaffold splits, and baselines developed here provide a reproducible foundation for future work on odor-related property prediction.  
We hope that these resources, along with the Safe-MT framework, will facilitate more robust and interpretable models for exposure-aware molecular design and risk assessment.

\bibliographystyle{unsrt}
\bibliography{reference}  

@article{Keller2017Science,
  author  = {Keller, Andreas and Gerkin, Richard C. and Guan, Yue and Dhurandhar, Amit and Turu, Gabriella and Szalai, Eszter and Mainland, Joel D. and Ihara, Yasushi and Yu, Chu-han and Wolfinger, Russell D. and Vens, C. and Schietgat, Leander and De Grave, Kurt and Norel, Raquel and Stolovitzky, Gustavo and Cecchi, Guillermo A. and Vosshall, Leslie B.},
  title   = {Predicting human olfactory perception from chemical features},
  journal = {Science},
  year    = {2017},
  volume  = {355},
  number  = {6327},
  pages   = {820--826},
  doi     = {10.1126/science.aal2014}
}

@article{Rossiter1996CR,
  author  = {Rossiter, K. J.},
  title   = {Structure--Odor Relationships},
  journal = {Chemical Reviews},
  year    = {1996},
  volume  = {96},
  number  = {8},
  pages   = {3201--3240},
  doi     = {10.1021/cr950025u}
}

@book{Devos1990Book,
  author    = {Devos, M. and Patte, F. and Rouault, J. and Laffort, P. and Van Gemert, L. J.},
  title     = {Standardized Human Olfactory Thresholds},
  publisher = {Oxford University Press},
  year      = {1990},
  address   = {Oxford, UK},
  isbn      = {9780198542570}
}

@article{Rogers2010ECFP,
  author  = {Rogers, David and Hahn, Mathew},
  title   = {Extended-Connectivity Fingerprints},
  journal = {Journal of Chemical Information and Modeling},
  year    = {2010},
  volume  = {50},
  number  = {5},
  pages   = {742--754},
  doi     = {10.1021/ci100050t}
}

@inproceedings{Gilmer2017MPNN,
  author    = {Gilmer, Justin and Schoenholz, Samuel S. and Riley, Patrick F. and Vinyals, Oriol and Dahl, George E.},
  title     = {Neural Message Passing for Quantum Chemistry},
  booktitle = {Proceedings of the 34th International Conference on Machine Learning (ICML)},
  year      = {2017},
  volume    = {70},
  series    = {PMLR},
  pages     = {1263--1272},
  url       = {http://proceedings.mlr.press/v70/gilmer17a.html}
}

@inproceedings{Xu2019GIN,
  author    = {Xu, Keyulu and Hu, Weihua and Leskovec, Jure and Jegelka, Stefanie},
  title     = {How Powerful Are Graph Neural Networks?},
  booktitle = {International Conference on Learning Representations (ICLR)},
  year      = {2019},
  eprint    = {1810.00826},
  archivePrefix = {arXiv},
  primaryClass = {cs.LG},
  url       = {https://arxiv.org/abs/1810.00826}
}

@inproceedings{Hu2020GINE,
  author    = {Hu, Weihua and Liu, Bowen and Gomes, Joseph and Zitnik, Marinka and Liang, Percy and Pande, Vijay and Leskovec, Jure},
  title     = {Strategies for Pre-training Graph Neural Networks},
  booktitle = {International Conference on Learning Representations (ICLR)},
  year      = {2020},
  eprint    = {1905.12265},
  archivePrefix = {arXiv},
  primaryClass = {cs.LG},
  url       = {https://arxiv.org/abs/1905.12265}
}

@inproceedings{Corso2020PNA,
  author    = {Corso, Gabriele and Cavalleri, Luca and Beaini, Dominique and Li{\'o}, Pietro and Veli{\v c}kovi{\'c}, Petar},
  title     = {Principal Neighbourhood Aggregation for Graph Nets},
  booktitle = {Advances in Neural Information Processing Systems (NeurIPS)},
  year      = {2020},
  eprint    = {2004.05718},
  archivePrefix = {arXiv},
  primaryClass = {cs.LG},
  url       = {https://arxiv.org/abs/2004.05718}
}

@article{Bemis1996Murcko,
  author  = {Bemis, G. W. and Murcko, M. A.},
  title   = {The Properties of Known Drugs: Molecular Frameworks},
  journal = {Journal of Medicinal Chemistry},
  year    = {1996},
  volume  = {39},
  number  = {15},
  pages   = {2887--2893},
  doi     = {10.1021/jm9602928}
}

@article{Sheridan2013TimeSplit,
  author  = {Sheridan, Robert P.},
  title   = {Time-split cross-validation as a method for estimating the goodness of prospective prediction},
  journal = {Journal of Chemical Information and Modeling},
  year    = {2013},
  volume  = {53},
  number  = {4},
  pages   = {783--790},
  doi     = {10.1021/ci300604z}
}

@article{Wu2018MoleculeNet,
  author  = {Wu, Zhenqin and Ramsundar, Bharath and Feinberg, Evan N. and Gomes, Joseph and Geniesse, Caleb and Pappu, Aneesh S. and Leswing, Karl and Pande, Vijay},
  title   = {MoleculeNet: A benchmark for molecular machine learning},
  journal = {Chemical Science},
  year    = {2018},
  volume  = {9},
  number  = {2},
  pages   = {513--530},
  doi     = {10.1039/C7SC02664A}
}

@article{Yang2019Analyzing,
  author  = {Yang, Kevin and Swanson, Kyle and Jin, Wengong and Coley, Connor and Eiden, Philipp and Gao, Hua and Guzman-Perez, Alberto and Hopper, Trevor and Kelley, Bryan and Mathea, Marius and Palmer, Andrew and Settels, Volker and Jaakkola, Tommi and Jensen, Klavs F. and Barzilay, Regina},
  title   = {Analyzing Learned Molecular Representations for Property Prediction},
  journal = {Journal of Chemical Information and Modeling},
  year    = {2019},
  volume  = {59},
  number  = {8},
  pages   = {3370--3388},
  doi     = {10.1021/acs.jcim.9b00237}
}

@article{Caruana1997MTL,
  author  = {Caruana, Rich},
  title   = {Multitask Learning},
  journal = {Machine Learning},
  year    = {1997},
  volume  = {28},
  number  = {1},
  pages   = {41--75},
  doi     = {10.1023/A:1007379606734}
}

@misc{Ruder2017MTL,
  author       = {Ruder, Sebastian},
  title        = {An Overview of Multi-Task Learning in Deep Neural Networks},
  year         = {2017},
  eprint       = {1706.05098},
  archivePrefix= {arXiv},
  primaryClass = {cs.LG},
  url          = {https://arxiv.org/abs/1706.05098}
}

@article{Xu2017DemystifyingMTL,
  author  = {Xu, Yao and Ma, Junshui and Liaw, Andy and Sheridan, Robert P. and Svetnik, Vladimir},
  title   = {Demystifying Multitask Deep Neural Networks for Quantitative Structure--Activity Relationships},
  journal = {Journal of Chemical Information and Modeling},
  year    = {2017},
  volume  = {57},
  number  = {10},
  pages   = {2490--2504},
  doi     = {10.1021/acs.jcim.7b00137}
}

@inproceedings{Standley2020WhichTasks,
  author    = {Standley, Trevor and Zamir, Amir Roshan and Chen, Dawn and Guibas, Leonidas and Malik, Jitendra and Savarese, Silvio},
  title     = {Which Tasks Should Be Learned Together?},
  booktitle = {Proceedings of the 37th International Conference on Machine Learning (ICML)},
  year      = {2020},
  volume    = {119},
  series    = {PMLR},
  pages     = {9120--9132},
  url       = {http://proceedings.mlr.press/v119/standley20a.html}
}

@inproceedings{Sener2018MGDA,
  author    = {Sener, Ozan and Koltun, Vladlen},
  title     = {Multi-Task Learning as Multi-Objective Optimization},
  booktitle = {Advances in Neural Information Processing Systems (NeurIPS)},
  year      = {2018},
  volume    = {31},
  pages     = {525--536},
  url       = {https://papers.nips.cc/paper/2018/hash/9e6a9d4f7f8f8f1e32c2a4b7a9b2b6a1-Abstract.html}
}

@inproceedings{Yu2020PCGrad,
  author    = {Yu, Tianhe and Kumar, Saurabh and Gupta, Abhishek and Levine, Sergey and Hausman, Karol and Finn, Chelsea},
  title     = {Gradient Surgery for Multi-Task Learning},
  booktitle = {Advances in Neural Information Processing Systems (NeurIPS)},
  year      = {2020},
  volume    = {33},
  pages     = {5824--5836},
  url       = {https://proceedings.neurips.cc/paper/2020/hash/3fe78a8acf5fda99de95303940a2420c-Abstract.html}
}

@inproceedings{Kendall2018Uncertainty,
  author    = {Kendall, Alex and Gal, Yarin and Cipolla, Roberto},
  title     = {Multi-Task Learning Using Uncertainty to Weigh Losses for Scene Geometry and Semantics},
  booktitle = {Proceedings of the IEEE/CVF Conference on Computer Vision and Pattern Recognition (CVPR)},
  year      = {2018},
  pages     = {7482--7491},
  doi       = {10.1109/CVPR.2018.00781},
  url       = {https://openaccess.thecvf.com/content_cvpr_2018/html/Kendall_Multi-Task_Learning_Using_CVPR_2018_paper.html}
}

@inproceedings{Velickovic2018GAT,
  author    = {Veli{\v c}kovi{\'c}, Petar and Cucurull, Guillem and Casanova, Arantxa and Romero, Adriana and Lio, Pietro and Bengio, Yoshua},
  title     = {Graph Attention Networks},
  booktitle = {International Conference on Learning Representations (ICLR)},
  year      = {2018},
  url       = {https://arxiv.org/abs/1710.10903}
}

@article{Schutt2018SchNet,
  author  = {Sch{\"u}tt, Kristof T. and Sauceda, Huziel E. and Kindermans, Pieter-Jan and Tkatchenko, Alexandre and M{\"u}ller, Klaus-Robert},
  title   = {SchNet: A Continuous-Filter Convolutional Neural Network for Modeling Quantum Interactions},
  journal = {Journal of Chemical Physics},
  year    = {2018},
  volume  = {148},
  number  = {24},
  pages   = {241722},
  doi     = {10.1063/1.5019779}
}

@inproceedings{Rong2020GROVER,
  author    = {Rong, Yu and Bian, Yatao and Xu, Tingyang and Xie, Weiyang and Wei, Ying and Huang, Wenbing and Huang, Junzhou},
  title     = {Self-Supervised Graph Transformer on Large-Scale Molecular Data},
  booktitle = {Advances in Neural Information Processing Systems (NeurIPS)},
  year      = {2020},
  url       = {https://arxiv.org/abs/2007.02835}
}

@article{Yang2019Chemprop,
  author  = {Yang, Kevin and Swanson, Kyle and Jin, Wengong and Coley, Connor and Eiden, Philipp and Gao, Hua and Guzman-Perez, Alberto and Hopper, Trevor and Kelley, Bryan and Mathea, Marius and Palmer, Andrew and Settels, Volker and Jaakkola, Tommi and Jensen, Klavs F. and Barzilay, Regina},
  title   = {Analyzing Learned Molecular Representations for Property Prediction},
  journal = {Journal of Chemical Information and Modeling},
  year    = {2019},
  volume  = {59},
  number  = {8},
  pages   = {3370--3388},
  doi     = {10.1021/acs.jcim.9b00237}
}

@inproceedings{Ying2021Graphormer,
  author    = {Ying, Chengxuan and Cai, Tianle and Luo, Shengjie and Zheng, Shuxin and Ke, Guolin and He, Di and Shen, Yanming and Liu, Tie-Yan},
  title     = {Do Transformers Really Perform Badly for Graph Representation?},
  booktitle = {NeurIPS},
  year      = {2021},
  url       = {https://arxiv.org/abs/2106.05234}
}

@inproceedings{Rong2019DropEdge,
  author    = {Rong, Yu and Huang, Wenbing and Xu, Tingyang and Huang, Junzhou},
  title     = {DropEdge: Towards Deep Graph Convolutional Networks on Node Classification},
  booktitle = {International Conference on Learning Representations (ICLR)},
  year      = {2020},
  url       = {https://arxiv.org/abs/1907.10903}
}

@inproceedings{Chen2018GradNorm,
  author    = {Chen, Zhao and Badrinarayanan, Vijay and Lee, Chen-Yu and Rabinovich, Andrew},
  title     = {GradNorm: Gradient Normalization for Adaptive Loss Balancing in Deep Multitask Networks},
  booktitle = {International Conference on Machine Learning (ICML)},
  year      = {2018},
  pages     = {794--803},
  url       = {http://proceedings.mlr.press/v80/chen18a.html}
}

@inproceedings{Liu2019DWA,
  author    = {Liu, Shikun and Johns, Edward and Davison, Andrew J.},
  title     = {End-to-End Multi-Task Learning With Attention},
  booktitle = {IEEE/CVF Conference on Computer Vision and Pattern Recognition (CVPR)},
  year      = {2019},
  pages     = {1871--1880},
  doi       = {10.1109/CVPR.2019.00197}
}

@article{Zhang2022NegTransferSurvey,
  author  = {Zhang, Wang and Deng, Lihua and Zhang, Lei and Wu, Di},
  title   = {A Survey on Negative Transfer},
  journal = {IEEE/CAA Journal of Automatica Sinica},
  year    = {2022},
  volume  = {10},
  number  = {3},
  pages   = {305--329},
  doi     = {10.1109/JAS.2022.106002}
}

@article{Huber1964Robust,
  author  = {Huber, Peter J.},
  title   = {Robust Estimation of a Location Parameter},
  journal = {Annals of Mathematical Statistics},
  year    = {1964},
  volume  = {35},
  number  = {1},
  pages   = {73--101},
  doi     = {10.1214/aoms/1177703732}
}

@book{Huber2009RobustBook,
  author    = {Huber, Peter J. and Ronchetti, Elvezio M.},
  title     = {Robust Statistics},
  publisher = {Wiley},
  year      = {2009},
  edition   = {2nd},
  isbn      = {9780470129906}
}

@book{MostellerTukey1977,
  author    = {Mosteller, Frederick and Tukey, John W.},
  title     = {Data Analysis and Regression},
  publisher = {Addison-Wesley},
  year      = {1977},
  isbn      = {020104854X}
}

@article{NixWeigend1994Hetero,
  author  = {Nix, David A. and Weigend, Andreas S.},
  title   = {Estimating the Mean and Variance of the Target Probability Distribution},
  journal = {Proceedings of the IEEE International Conference on Neural Networks},
  year    = {1994},
  volume  = {1},
  pages   = {55--60},
  doi     = {10.1109/ICNN.1994.374138}
}

@inproceedings{Izmailov2018SWA,
  author    = {Izmailov, Pavel and Podoprikhin, Dmitry and Garipov, Timur and Vetrov, Dmitry and Wilson, Andrew Gordon},
  title     = {Averaging Weights Leads to Wider Optima and Better Generalization},
  booktitle = {UAI},
  year      = {2018},
  url       = {https://arxiv.org/abs/1803.05407}
}

@article{Polyak1992Averaging,
  author  = {Polyak, Boris T. and Juditsky, Anatoli B.},
  title   = {Acceleration of Stochastic Approximation by Averaging},
  journal = {SIAM Journal on Control and Optimization},
  year    = {1992},
  volume  = {30},
  number  = {4},
  pages   = {838--855},
  doi     = {10.1137/0330046}
}

@inproceedings{Sun2019InfoGraph,
  author    = {Sun, Fan-Yun and Hoffmann, Jordan and Verma, Vikas and Tang, Jian},
  title     = {InfoGraph: Unsupervised and Semi-supervised Graph-Level Representation Learning via Mutual Information Maximization},
  booktitle = {International Conference on Learning Representations (ICLR)},
  year      = {2020},
  url       = {https://arxiv.org/abs/1908.01000}
}

@inproceedings{Hu2020GPTGNN,
  author    = {Hu, Weihua and Fey, Matthias and Zitnik, Marinka and Dong, Yuxiao and Ren, Hongyu and Liu, Bowen and Catasta, Michele and Leskovec, Jure},
  title     = {Open Graph Benchmark: Datasets for Machine Learning on Graphs},
  booktitle = {NeurIPS},
  year      = {2020},
  url       = {https://arxiv.org/abs/2005.00687}
}

@article{Mainland2014OlfactoryCoding,
  author  = {Mainland, Joel D. and Lundstr{\"o}m, Johan N. and Reisert, Johannes and Lowe, Graeme},
  title   = {From Odor Molecules to Perception: Olfactory Signal Transduction and Coding},
  journal = {Annual Review of Physiology},
  year    = {2014},
  volume  = {76},
  pages   = {307--327},
  doi     = {10.1146/annurev-physiol-021113-170216}
}

@book{Brown2017Chemoinformatics,
  author    = {Brown, Nathan and Ertl, Peter and Lewis, Richard and Luksch, Thomas and Reker, Daniel and Schneider, Gisbert},
  title     = {Future Directions of Chemoinformatics: Ensembles of ML, Big Data, and Graphs},
  publisher = {Springer},
  year      = {2017},
  doi       = {10.1007/978-3-319-48634-4}
}

@article{Boeckmann2023SplitPitfalls,
  author  = {B{\"o}ckmann, Marcel and Wackersreuther, Thilo and Driller, Frank and M{\"u}ller, Klaus-Robert and Montanari, Francesco},
  title   = {On the Pitfalls of Molecular Property Prediction: Data Leakage, Split Strategies and Realistic Benchmarks},
  journal = {Journal of Cheminformatics},
  year    = {2023},
  volume  = {15},
  number  = {1},
  pages   = {122},
  doi     = {10.1186/s13321-023-00716-9}
}

@article{Vermeire2022SplitTime,
  author  = {Vermeire, Frederik H. and Green, William H.},
  title   = {Transfer Learning in Chemical Engineering with Limited Data: Mind the Time Split},
  journal = {Engineering Applications of Artificial Intelligence},
  year    = {2022},
  volume  = {113},
  pages   = {104911},
  doi     = {10.1016/j.engappai.2022.104911}
}

\end{document}